\documentclass[10pt,twocolumn,letterpaper]{article}

\usepackage{cvpr}              % To produce the CAMERA-READY version
\usepackage{graphicx}
\usepackage{amsmath}
\usepackage{amssymb}
\usepackage{booktabs}
\usepackage{array}
\usepackage{relsize}
\usepackage{multirow}

\makeatletter
\@namedef{ver@everyshi.sty}{}
\makeatother
 \usepackage{tikz,pgfplots}
\usetikzlibrary{positioning}

\usepackage[pagebackref,breaklinks,colorlinks]{hyperref}

\newcommand\imagew{0.45\columnwidth}
\newcommand\imageh{0.3\columnwidth}
\usepackage[capitalize]{cleveref}
\crefname{section}{Sec.}{Secs.}
\Crefname{section}{Section}{Sections}
\Crefname{table}{Table}{Tables}
\crefname{table}{Tab.}{Tabs.}

\newcommand*\samethanks[1][\value{footnote}]{\footnotemark[#1]}

\def\cvprPaperID{4887} % *** Enter the CVPR Paper ID here
\def\confName{CVPR}
\def\confYear{2022}

\begin{document}

\title {ClipCap: CLIP Prefix for Image Captioning}

\author{Ron Mokady\thanks{Equal contribution.} \qquad Amir Hertz\samethanks \qquad Amit H. Bermano \\ 
The Blavatnik School of Computer Science, Tel Aviv University\\
}
% \and

\maketitle
% \twocolumn[{
% \renewcommand\twocolumn[1][]{#1}
% \maketitle
%   \input{images/teaser_horizontal}
% }]

%%%%%%%%% ABSTRACT
\begin{abstract}

Image captioning is a fundamental task in vision-language understanding, where the model predicts a textual informative caption to a given input image. In this paper, we present a simple approach to address this task. We use CLIP encoding as a prefix to the caption, by employing a simple mapping network, and then fine-tunes a language model to generate the image captions. The recently proposed CLIP model contains rich semantic features which were trained with textual context, making it best for vision-language perception. Our key idea is that together with a pre-trained language model (GPT2), we obtain a wide understanding of both visual and textual data. Hence, our approach only requires rather quick training to produce a competent captioning model. Without additional annotations or pre-training, it efficiently generates meaningful captions for large-scale and diverse datasets. Surprisingly, our method works well even when only the mapping network is trained, while both CLIP and the language model remain frozen, allowing a lighter architecture with less trainable parameters. Through quantitative evaluation, we demonstrate our model achieves comparable results to state-of-the-art methods on the challenging Conceptual Captions and nocaps datasets, while it is simpler, faster, and lighter. Our code is available in \url{https://github.com/rmokady/CLIP_prefix_caption}.

\end{abstract}

\renewcommand{\arraystretch}{1}
\setlength{\tabcolsep}{3pt}
\begin{figure}
  \centering
  \footnotesize
\begin{tabular}{p{\imagew} p{\imagew}}
\\
\includegraphics[width=\imagew,height=\imageh]{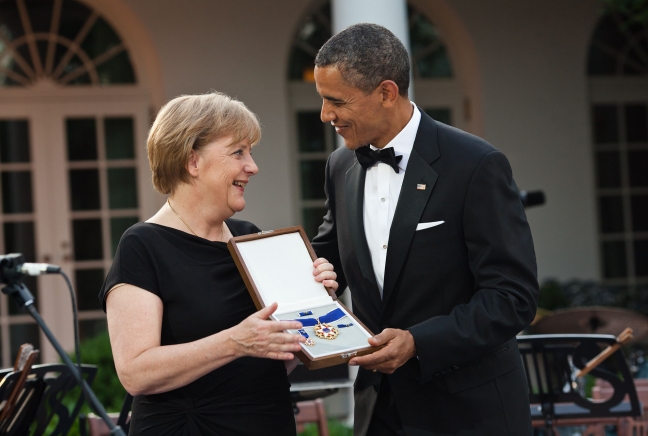} &
\includegraphics[width=\imagew,height=\imageh]{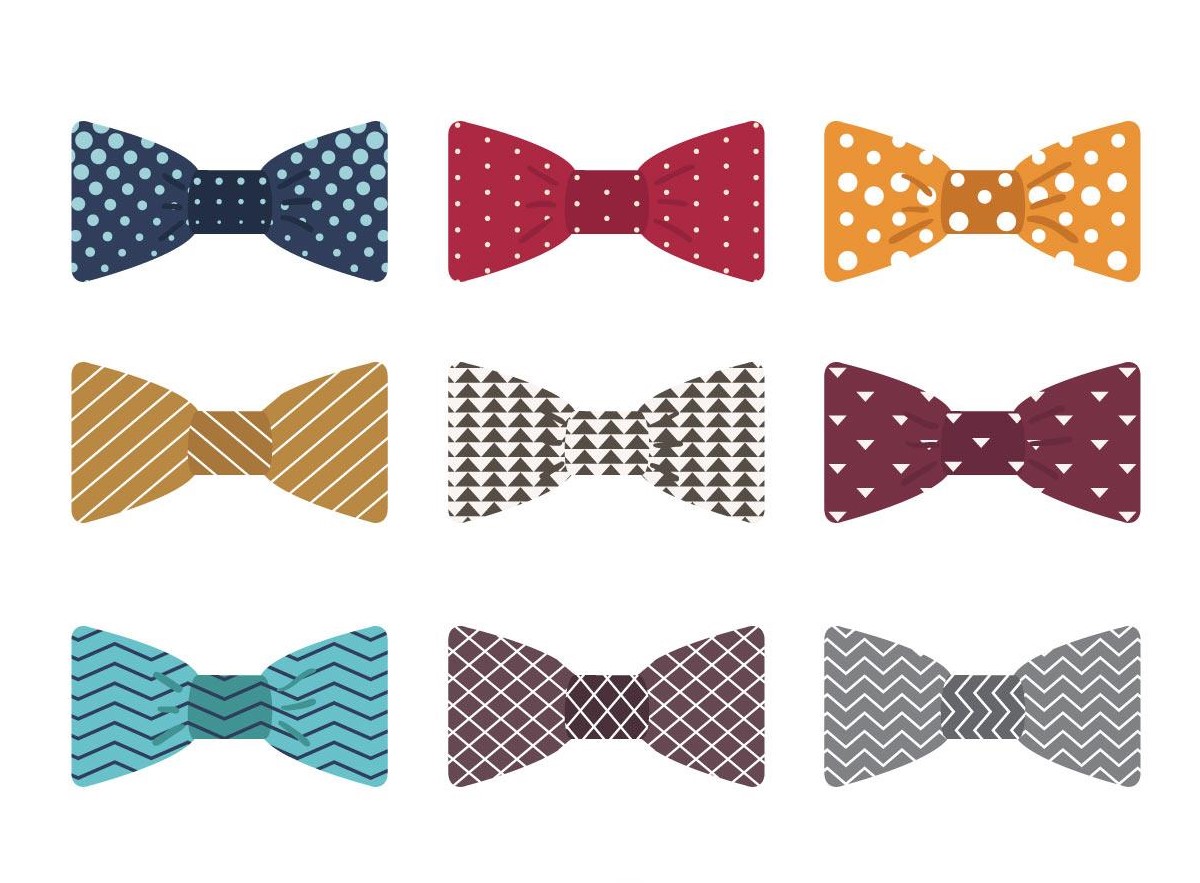} \\
A politician receives a gift from politician. & A collage of different colored ties on a white background. \\
\\
\includegraphics[width=\imagew,height=\imageh]{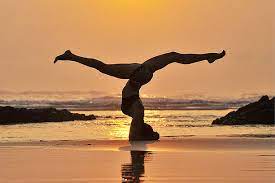} &
\includegraphics[width=\imagew,height=\imageh]{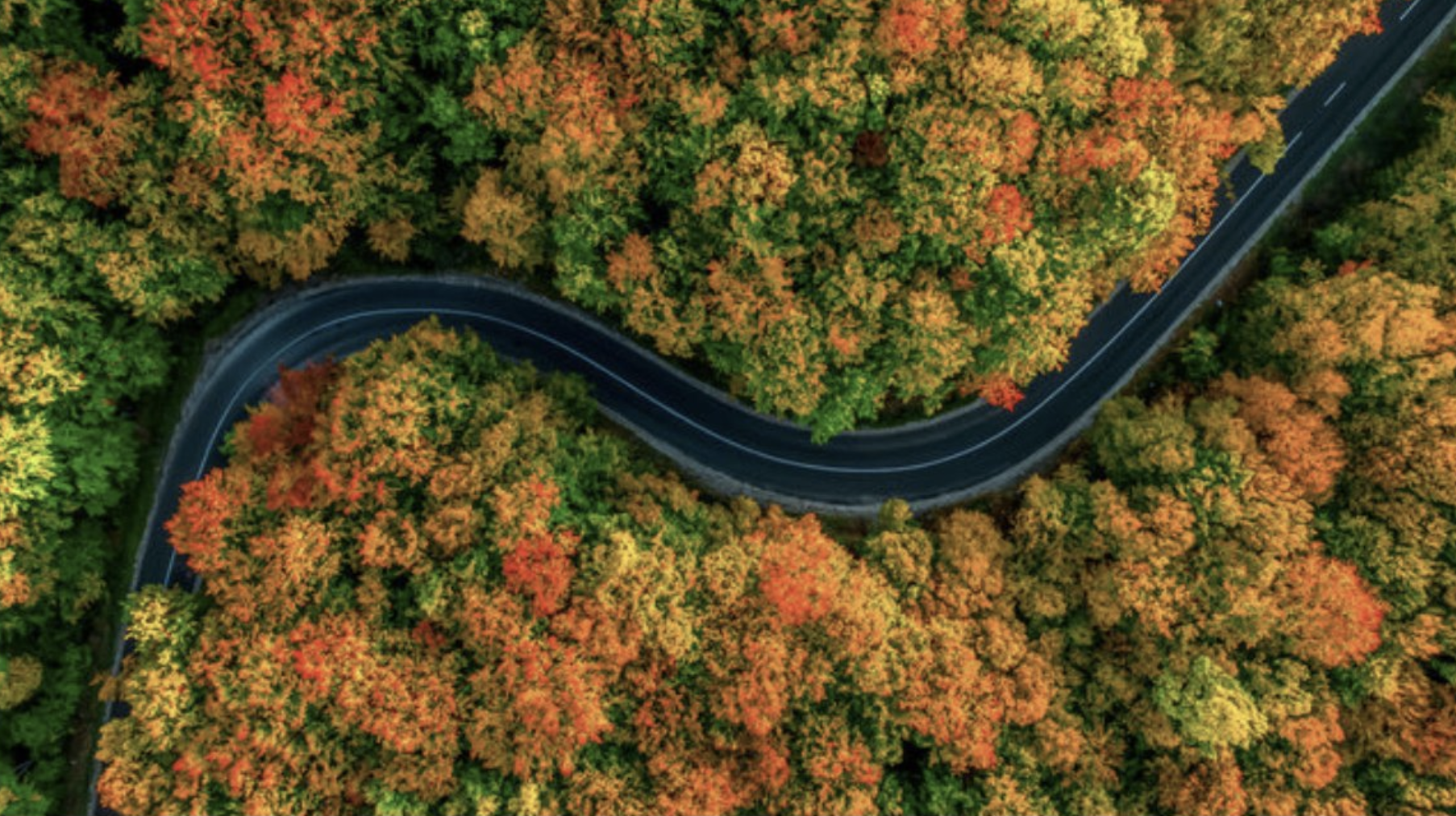} \\
Silhouette of a woman practicing yoga on the beach at sunset. & Aerial view of a road in autumn. 
\end{tabular}
%\end{center}
\vspace{-0.1cm}
\captionof{figure}{
Our ClipCap model produces captions depcting the respective images. Here, the results are of a model that was trained over the Conceptual Captions dataset.}
\vspace{-0.05cm}
\label{fig:teaser} 
\end{figure}

\section{Introduction}

\begin{figure*}

    \includegraphics[width=\textwidth, trim=0 0 0 0,clip]{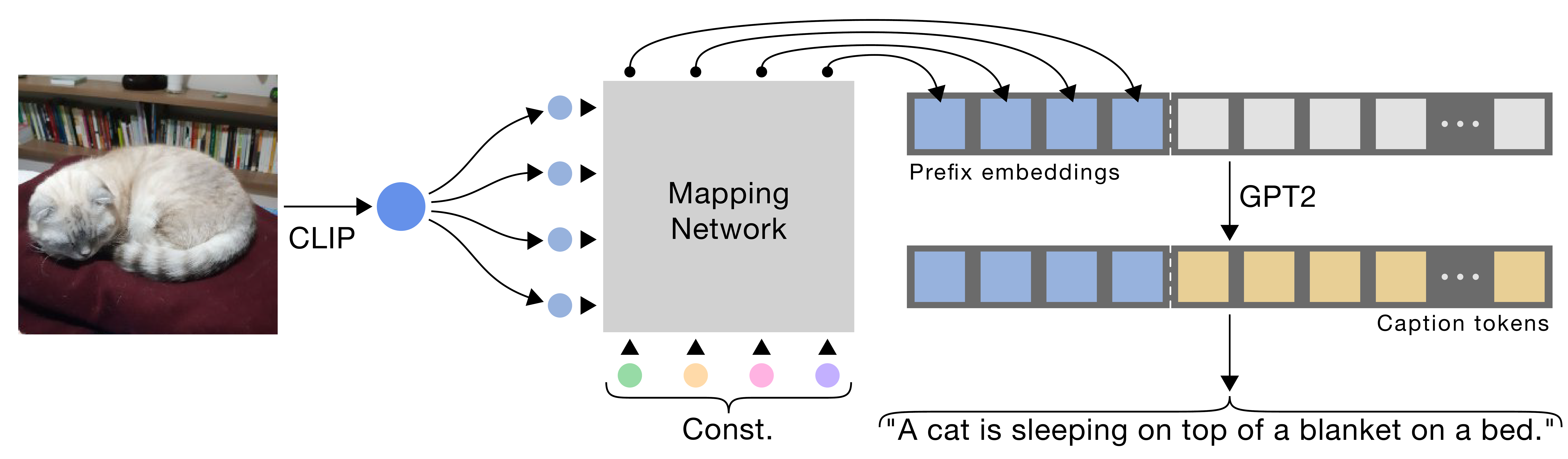}
     \vspace{-0.6cm}
    \caption{    %To extract a fixed length prefix, we train a lightweight mapping network from the CLIP embedding space to a frozen GPT-2.
    Overview of our transformer-based architecture, enabling the generation of meaningful captions while both CLIP and the language model, GPT-$2$, are frozen. To extract a fixed length prefix, we train a lightweight transformer-based mapping network from the CLIP embedding space and a learned constant to GPT-$2$. At inference, we employ GPT-$2$ to generate the caption given the prefix embeddings.
    %We utilize the transformer architecture for our mapping network, where learned constants are used as additional inputs. 
    We also suggest a MLP-based architecture, refer to Sec.~\ref{sec:method} for more details.
    }
    \vspace{-0.2cm}
    % Method overview. Our light weight image-captioning training includes small "Prefix Extraction" MLP and fine tuning of a pretrained GPT2-large model. \ron{RON: I'm kind of missing the mapping network which is important component. Also, how about making it double column? we have plenty of space.}
    \label{fig:method_diagram}
\end{figure*}

In image captioning, the task is to provide a meaningful and valid caption for a given input image in a natural language. 
This task poses two main challenges. The first is semantic understanding. This aspect ranges from simple tasks such as detecting the main object,
to more involved ones, such as understanding the relations between depicted parts of the image. For example, in the top-left image of Fig.~\ref{fig:teaser}, the model understands that the object is a gift. The second challenge is the large number of possible ways to describe a single image. In this aspect, the training dataset typically dictates the preferable option for a given image.

Many approaches have been proposed for image captioning  ~\cite{chen2014learning, fang2015captions, xu2015show, yao2018exploring, anderson2018bottom, tan2019lxmert, zhou2020unified, li2020oscar, stefanini2021show}. Typically, these works utilize an encoder for visual cues and a textual decoder to produce the final caption. Essentially, this induces the need to bridge the challenging gap between the visual and textual representations. For this reason, such models are resource hungry. They require extensive training time, a large number of trainable parameters, a massive dataset, and in some cases even additional annotations (such as detection results), which limit their practical applicability. % and ease of use. 

Excessive training time is even more restrictive for applications that require several training procedures. For instance, training multiple captioning models over various datasets could provide different users (or applications) with different captions for the same image. Additionally, given fresh samples, it is desirable to update the model routinely with the new data. Therefore, a \textit{lightweight} captioning model is preferable. Specifically, a model with faster training times and fewer trainable parameters would be beneficial, especially if it does not require additional supervision.

In this paper, we leverage powerful vision-language pre-trained models to simplify the captioning process.
More specifically, we use the  CLIP (Contrastive Language-Image Pre-Training) encoder, recently introduced by Radford et al.~\cite{radford2021learning}.  CLIP is designed to impose a shared representation for both images and text prompts. It is trained over a vast number of images and textual descriptions using a contrastive loss. Hence, its visual and textual representations are well correlated. As we demonstrate, this correlation saves training time and data requirements.

As illustrated in Fig.~\ref{fig:method_diagram}, our method produces a prefix for each caption by applying a mapping network over the CLIP embedding. This prefix is a fixed size embeddings sequence, concatenated to the caption embeddings. These are fed to a language model, which is fine-tuned along with the mapping network training. 
At inference, the language model generates the caption word after word, starting from the CLIP prefix. 
This scheme narrows the aforementioned gap between the visual and textual worlds, allowing the employment of a simple mapping network. 
To achieve even a lighter model, we introduce another variant of our method, where we train only the mapping network, while both CLIP and the language model are kept frozen. By utilizing the expressive transformer architecture, we successfully produce meaningful captions, while imposing substantially less trainable parameters. Our approach is inspired by Li et al.~\cite{li2021prefix}, which demonstrates the ability to efficiently adapt a language model for new tasks by concatenating a learned prefix. 
We use GPT-$2$ \cite{radford2019language} as our language model, which has been demonstrated to generate rich and diverse texts.

As our approach exploits the rich visual-textual representation of CLIP, our model requires significantly lower training time. For instance, we train our model on a single Nvidia GTX$1080$ GPU for $80$ hours over the three million samples of the massive Conceptual Captions dataset. Nevertheless, our model generalizes well to complex scenes, as can be seen in Fig.~\ref{fig:teaser} (e.g., practicing yoga on the beach at sunset).
We evaluate our method extensively, demonstrating successful realistic and meaningful captions.
Even though our model requires less training time, it still achieves comparable results to state-of-the-art approaches over the challenging Conceptual Captions \cite{sharma2018conceptual} and nocaps \cite{agrawal2019nocaps} datasets, and marginally lower for the more restricted COCO \cite{lin2014microsoft, chen2015microsoft} benchmark. 
In addition, we provide a thorough analysis of the required prefix length and the effect of fine-tuning the language model, including interpretation of our produced prefixes. 
Overall, our main contributions are as follow:
\begin{itemize}
    \item
A lightweight captioning approach that utilizes pre-trained frozen models for both visual and textual processing.
\item
   Even when the language model is fine-tuned, our approach is simpler and faster to train, while demonstrating comparable results to state-of-the-art over challenging datasets. 

\end{itemize}

\section{Related Works}
\label{sec:rw}

Recently, Radford et al.~\cite{radford2021learning} presented a novel approach, known as CLIP, to jointly represent images and text descriptions. CLIP comprises two encoders, one for visual cues and one for text. It was trained over more than $400$ million image-text pairs guided by unsupervised contrastive loss, resulting in rich semantic latent space shared by both visual and textual data. Many works have already used CLIP successfully for computer vision tasks that require the understanding of some auxiliary text, such as generating or editing an image based on a natural language condition \cite{patashnik2021styleclip, bau2021paint, gal2021stylegan}. 
In this paper, we utilize the powerful CLIP model for the task of image captioning.
Note that our method does not employ the CLIP's textual encoder, since there is no input text, and the output text is generated by a language model.

Commonly, image captioning \cite{stefanini2021show} models first encode the input pixels as feature vectors, which are then used to produce the final sequence of words. Early works utilize the features extracted from a pre-trained classification network \cite{chen2014learning, fang2015captions, xu2015show, chen2017sca}, 
while later works \cite{anderson2018bottom, zhou2020unified, li2020oscar} exploit the more expressive features of an object detection network \cite{ren2015faster}.
Though a pre-trained object detection network is available for the popular COCO benchmark \cite{lin2014microsoft, chen2015microsoft}, it is not necessarily true for other datasets. This implies that most methods would require additional object detection annotations to operate over new and diverse datasets. To further leverage the visual cues, an attention mechanism is usually utilized \cite{xu2015show, chen2017sca, anderson2018bottom} to focus on specific visual features.
Moreover, recent models apply self-attention \cite{yang2019learning, herdade2019image} or use an expressive visual Transformer \cite{dosovitskiy2020image} as an encoder \cite{liu2021cptr}. 
Our work uses the expressive embedding of CLIP for visual representation. Since CLIP was trained over an extremely large number of images, we can operate on any set of natural images without additional annotations.

To produce the caption itself, a textual decoder is employed. 
Early works have used LSTM variants
\cite{vinyals2015show, wang2017skeleton, chen2018regularizing}, while recent works \cite{herdade2019image, luo2021dual} adopted the improved transformer architecture  \cite{vaswani2017attention}. Built upon the transformer, one of the most notable works is BERT \cite{devlin2018bert}, demonstrating the dominance of the newly introduced paradigm. With this paradigm, the language model is first pre-trained over a large data collection to solve an auxiliary task. 
Then, the model is fine-tuned for a specific task, where additional supervision is used.
As our visual information resides in the prefix, we utilize a powerful auto-regressive language model, GPT-$2$ \cite{radford2019language}.
Considering the training loss term, earlier works adopt the effective cross-entropy, while contemporary methods also apply self-critical sequence training \cite{rennie2017self, zhang2017actor, gao2019self}. That is, an additional training stage to optimize the CIDEr metric. We deliberately refrain from this optimization to retain a quick training procedure.

Most close to ours, are works that employ vision-and-language pre-training to create a shared latent space of both vision and text \cite{tan2019lxmert, lu2019vilbert, li2020oscar, zhou2020unified, zhang2021vinvl}. Zhou et al.~\cite{zhou2020unified} use visual tokens extracted from object detector as a prefix to caption tokens. The entire model is then pre-trained to perform prediction utilizing the BERT \cite{devlin2018bert} architecture. Li et al.~\cite{li2020oscar} and Zhang et al.~\cite{zhang2021vinvl} also utilize BERT, but require the additional supervision of object tags. Hence, these methods are limited to datasets in which such object detectors or annotations are available. The approach of Wang et al.~\cite{wang2021simvlm} mitigate the need for supplementary annotations, but still perform an extensive pre-train process with millions of image-text pairs, resulting in a lengthy training time. 
This exhaustive pre-training step is required to compensate for the lack of joint representation of language and vision, which we inherently obtained by employing CLIP.

\section{Method}
\label{sec:method}

%\subsection{Problem Statement}
We start with our problem statement. Given a dataset of paired images and captions $\{x^i, c^i\}_{i=1}^{N}$, our goal is to learn the generation of a meaningful caption for an unseen input image. We can refer to the captions as a sequence of tokens $c^i = c^i_1, \ldots, c^i_\ell$, where we pad the tokens to a maximal length $\ell$. Our training objective is then the following:

\begin{align}
\max_\theta \sum_{i=1}^{N} \log p_\theta (c^i_1, \ldots, c^i_\ell | \: x^i),
\end{align}
where $\theta$ denotes the model's trainable parameters. 
Our key idea is to use the rich semantic embedding of CLIP, which contains, virtually, the essential visual data, as a condition. Following recent works \cite{zhou2020unified}, we consider the condition as a prefix to the caption. 
Since the required semantic information is encapsulated in the prefix, we can utilize an autoregressive language model that predicts the next token without considering future tokens. Thus, our objective can be described as:

\begin{align}
\max_\theta \sum_{i=1}^{N}\sum_{j=1}^{\ell} \log p_\theta (c^i_j | \: x^i, c^i_1, \ldots, c^i_{j-1})
\end{align}

\subsection{Overview}
An illustration of our method is provided in Fig.~\ref{fig:method_diagram}. We use GPT-$2$ (large) as our language model, and utilize its tokenizer to project the caption to a sequence of embeddings. To extract visual information from an image $x^i$, we use the visual encoder of a pre-trained CLIP \cite{radford2021learning} model.
Next, we employ a light mapping network, denoted $F$, to map the CLIP embedding to $k$ embedding vectors:
\begin{align}
p^i_1, \ldots, p^i_k = F(\text{CLIP}(x^i)).
\end{align}
Where each vector $p^i_j$ has the same dimension as a word embedding. 
We then concatenate the obtained visual embedding to the caption $c^i$ embeddings:
\begin{align}
Z^i = p^i_1, \ldots, p^i_k, c^i_1, \ldots, c^i_\ell.
\end{align}
During training, we feed the language model with the prefix-caption concatenation $\{Z^i\}_{i=1}^{N}$.
Our training objective is predicting the caption tokens conditioned on the prefix in an autoregressive fashion. 
To this purpose, we train the mapping component $F$ using the simple, yet effective, cross-entropy loss:
\begin{align}
\mathcal{L}_{X} = -\sum_{i=1}^{N}\sum_{j=1}^{\ell} \log p_\theta (c^i_j | \: p^i_1, \ldots, p^i_k, c^i_1, \ldots, c^i_{j-1}).
\end{align}
We now turn to discuss two variants of our method regarding the additional fine-tuning of the language model and their implications.

\subsection{Language model fine-tuning}
Our main challenge during training is to translate between the representations of CLIP and the language model. Even though both models develop a rich and diverse representation of text, their latent spaces are independent, as they were not jointly trained. Moreover, each captioning dataset incorporates a different style, which may not be natural for the pre-trained language model. Hence, we propose fine-tuning the language model during the training of the mapping network. This provides additional flexibility for the networks and yields a more expressive outcome.

However, fine-tuning the language model naturally increases the number of trainable parameters substantially. Thus, we present an additional variant of our approach, in which we keep the language model fixed during training. Our attempt to adjust a frozen language model is inspired by the work of Li and Liang \cite{li2021prefix}. In their work, they accommodate such a pre-trained model to an unfamiliar task by learning only a prefix. Such prefix is automatically optimized to steer the language model towards the new objective during a standard training procedure. Following this approach, we suggest avoiding the fine-tuning to realize an even lighter model, where only the mapping network is trained. As presented in Section \ref{sec:res}, our model not only produces realistic and meaningful captions, but also achieves superior results for some of the experiments without fine-tuning the language model. Note that fine-tuning CLIP does not benefit resulting quality, but does increase training time and complexity. We hence postulate that the CLIP space already encapsulates the required information, and adapting it towards specific styles does not contribute to flexibility.

\subsection{Mapping Network Architecture}

 Our key component is the mapping network, which translates the CLIP embedding to the GPT-$2$ space.   
When the language model is simultaneously fine-tuned, the mapping is less challenging, as we easily control both networks. Therefore, in this case, we can employ a simple Multi-Layer Perceptron (MLP). We have achieved realistic and meaningful captions even when utilizing only a single hidden layer, as CLIP is pre-trained for a vision-language objective.

Nevertheless, when the language model is frozen, we propose utilizing the more expressive transformer \cite{vaswani2017attention} architecture. The transformer enables global attention between input tokens while reducing the number of parameters for long sequences. This allows us to improve our results by increasing prefix size, as shown in Section.~\ref{sec:res}. We feed the transformer network with two inputs, the visual encoding of CLIP and a learned constant input. The constant has a dual role, first, to retrieve meaningful information from CLIP embedding through the multi-head attention. Second, it learns to adjust the fixed language model to the new data. 
This is demonstrated in Section.~\ref{sec:res}, where we offer interpretability for our generated prefix. As can be seen, when the language model is fixed, the transformer mapping network learns a meticulous set of embeddings without any textual meaning. These are optimized to tame the language model.

\subsection{Inference}
During inference, we extract the visual prefix of an input image $x$ using the CLIP encoder and the mapping network $F$.
We start generating the caption conditioned on the visual prefix, and predict the next tokens one by one, guided by the language model output. For each token, the language model outputs probabilities for all vocabulary tokens, which are used to determine the next one by employing a greedy approach or beam search.

\begin{table*}
\begin{center}

$(A)$ \textbf{Conceptual Captions}
\begin{tabular}{lccccc}
\toprule
Model &  ROUGE-L $\uparrow$ & CIDEr $\uparrow$ & SPICE $\uparrow$ & \#Params (M) $\downarrow$ & Training Time  $\downarrow$ \\ 
\bottomrule
\midrule
VLP & $24.35$ & $ 77.57$ & $16.59$ & 115 & $ 1200$h (V$100$)\\
\toprule
Ours; MLP + GPT2 tuning & $\mathbf{26.71}$ & $\mathbf{87.26}$ & $\mathbf{18.5}$ & $156$ & $80$h (GTX$1080$)\\
\midrule
% Ours w/ beam search & $26.51$ & $89.58$ & $18.2$ & 31 & $ 168$ (GTX$1080$)\\
% \midrule
Ours; Transformer & $25.12$ & $71.82$ & $16.07$ & $\mathbf{43}$ & $
\mathbf{72}$\textbf{h} (GTX$1080$) \\
\bottomrule 
\end{tabular}
% {'Bleu_1': 0.24489445029532422, 'Bleu_2': 0.1429134077174941, 'Bleu_3': 0.09088275309299156, 'Bleu_4': 0.060227730730366164, 'METEOR': 0.10208756150946528, 'ROUGE_L': 0.24821681573999405, 'CIDEr': 0.7063327299389974, 'SPICE': 0.16029866781695395}
\vspace{0.15cm}

$(B)$ \textbf{nocaps}

\begin{tabular}{p{0.17\textwidth}|cc|cc|cc|cc|cc}
\toprule

& \multicolumn{2}{c}{in-domain} & \multicolumn{2}{c}{near-domain} & \multicolumn{2}{c}{out-of-domain} & \multicolumn{2}{c}{Overall} & & \\
Model &  CIDEr$\uparrow$   & SPICE $\uparrow$  & CIDEr & SPICE & CIDEr & SPICE & CIDEr & SPICE & Params$\downarrow$ & Time$\downarrow$  \\ 
\bottomrule
\midrule
BUTD ~\cite{anderson2018bottom}  & $74.3$ & $ 11.5$ & $56.9$ & $10.3$ & $30.1$ & $8.1$ & $ 54.3$ & $10.1$ & $52$ & $960$h  \\
\midrule
Oscar ~\cite{li2020oscar} & $79.6$ & $ \mathbf{12.3}$ & $66.1$ & $\mathbf{11.5}$ & $45.3$ & $\mathbf{9.7}$ & $ 63.8$ & $\mathbf{11.2}$ & $135$ & $74$h\\
\toprule
Ours; MLP + GPT2 tuning & $79.73$ & $ 12.2$ & $\mathbf{67.69}$ & $11.26$ & $\mathbf{49.35}$ & $ \mathbf{9.7}$ & $ 65.7$ & $11.1$& $156$ & $7$h\\
\midrule
Ours; Transformer & $\mathbf{84.85}$ & $ 12.14$ & $66.82$ & $10.92$ & $49.14$ & $9.57$ & $ \mathbf{65.83}$ & $10.86$ &$\mathbf{43}$ & $\mathbf{6}$\textbf{h} \\
% \midrule
% Ours & $26.71$ & $87.26$ & $18.5$ & 31 & $ 168$ (GTX$1080$)\\
% \midrule
\midrule
\end{tabular}

\vspace{0.15cm}
$(C)$ \textbf{COCO}

\begin{tabular}{p{0.26\textwidth}cccccc}
\toprule
Model &   B@$4\uparrow$  & METEOR $\uparrow$ & CIDEr $\uparrow$ & SPICE $\uparrow$ & \#Params (M) $\downarrow$ & Training Time $\downarrow$ \\ 
\bottomrule
\midrule
BUTD ~\cite{anderson2018bottom} &  $36.2$ & $27.0$ & $113.5$ & $20.3$ & 52 & $960$h (M$40$)\\
\midrule
VLP ~\cite{zhou2020unified} &  $36.5$ & $28.4$ & $117.7$ & $21.3$ & 115 & $48$h (V$100$)\\
\midrule
Oscar ~\cite{li2020oscar} & $\mathbf{36.58}$ & $\mathbf{30.4}$ &  $\mathbf{124.12}$ & $\mathbf{23.17}$ &  $135$ & $74$h (V$100$)\\

\toprule
Ours; Transformer   & $33.53$ & $27.45$ &  $113.08$ & $21.05$ & $\mathbf{43}$ & $\mathbf{6}$\textbf{h} (GTX$1080$)\\
\midrule

Ours; MLP + GPT2 tuning & $32.15$ & $27.1$ &  $108.35$ & $20.12$ & $156$ & $7$h (GTX$1080$)\\

\midrule

\multicolumn{7}{c}{$(D)$ \textbf{Ablation}} \\
\bottomrule
\midrule

Ours; Transformer + GPT2 tuning  & $32.22$ & $27.79$ &  $109.83$ & $20.63$ & $167$ & $7$h (GTX$1080$) \\

\midrule
Ours; MLP & $27.39$ & $24.4$ & $92.38$ & $18.04$ & $32$ & $6$h (GTX$1080$)\\
\midrule

\end{tabular}

\end{center}
\vspace{-0.1cm}
\caption{Quantitative evaluation. As can be seen, our method achieves comparable results for both nocaps and Conceptual Captions with much faster training time. %*Training time of VLP and Oscar does not include pre-training
}
%\vspace{-0.3cm}
\label{tab:num} 
\end{table*}

\begin{figure*}
\footnotesize
\centering
\begin{tabular}{p{0.2\columnwidth} p{0.33\columnwidth}p{0.33\columnwidth}p{0.33\columnwidth}p{0.33\columnwidth}p{0.32\columnwidth}}
&
\includegraphics[trim={2cm 0 0 0},clip,width=0.34\columnwidth,height=0.22\columnwidth]{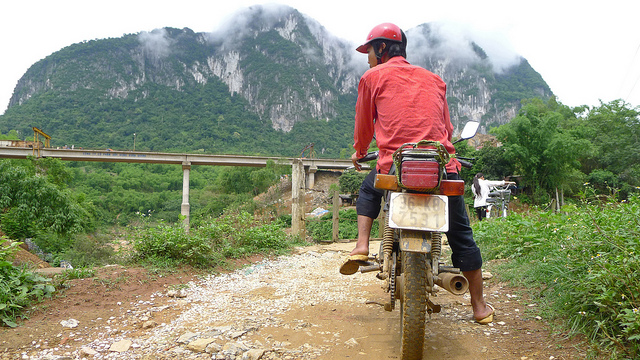} &
\includegraphics[width=0.34\columnwidth,height=0.22\columnwidth]{images/COCO_val2014_000000060623.jpg} &
\includegraphics[trim={0, .1cm 0 .2cm},clip,width=0.34\columnwidth,height=0.22\columnwidth]{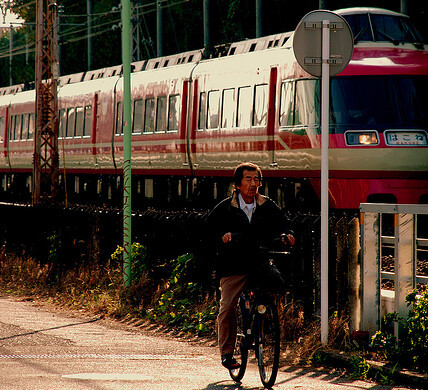} &
\includegraphics[trim={0 10cm 0 2cm},clip,width=0.34\columnwidth,height=0.22\columnwidth]{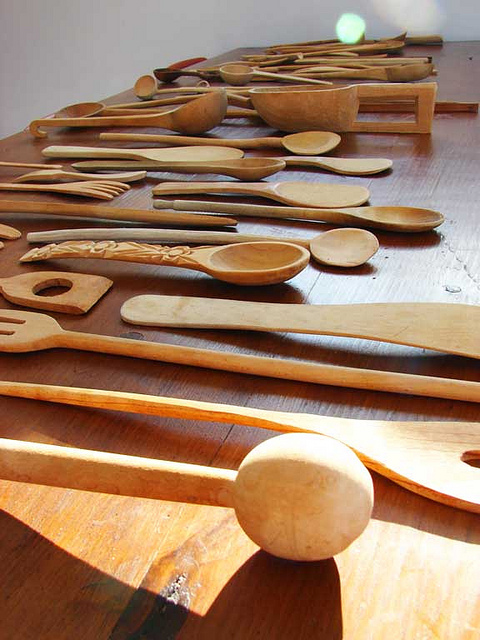} &
\includegraphics[trim={0 1cm 0 1cm},clip,width=0.34\columnwidth,height=0.22\columnwidth]{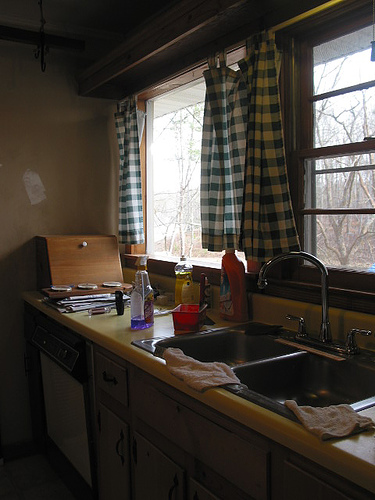} \\
% \hline
\begin{tabular}{l}Ground Truth \end{tabular}& A man with a red helmet on a small moped on a dirt road & A young girl inhales with the intent of blowing out a candle.  
& A man on a bicycle riding next to a train. & a wooden cutting board topped with sliced up food.  & A kitchen is shown with a variety of items on the counters.\\
\hline
\begin{tabular}{l}Oscar \end{tabular}& a man riding a motorcycle down a dirt road. & a woman sitting at a table with a plate of food.  & a woman riding a bike down a street next to a train. & a woman sitting at a table with a plate of food.  & a kitchen with a sink, dishwasher and a window.\\
\hline
\begin{tabular}{l}
Ours; MLP + \\ GPT2 tuning
\end{tabular}& a man riding a motorcycle on a dirt road. & a woman is eating a piece of cake with a candle. & a man is standing next to a train. & a row of wooden cutting boards with wooden spoons. & a kitchen with a sink, stove, and window.\\
\hline
\begin{tabular}{l}
Ours; \\ Transformer
\end{tabular}  & a man is riding a motorbike on a dirt road. & a young girl sitting at a table with a cup of cake. &  a man is standing next to a train.  & a wooden table with a bunch of wood tools on it.   & a kitchen with a sink and a window. \\

\end{tabular}
\vspace{-0.2cm}
\caption{Uncurated results of the first five images in the COCO test set (Karpathy et al.~\cite{karpathy2015deep} split).}% \amirc{uncomment above for one column option}}
\vspace{-0.2cm}
\label{fig:img} 
\end{figure*}

\begin{figure*}
\footnotesize
\centering
\begin{tabular}{p{0.2\columnwidth} p{0.33\columnwidth}p{0.33\columnwidth}p{0.33\columnwidth}p{0.33\columnwidth}p{0.33\columnwidth}}
&
\includegraphics[trim={0 3cm 0 2cm},clip,width=0.34\columnwidth,height=0.24\columnwidth]{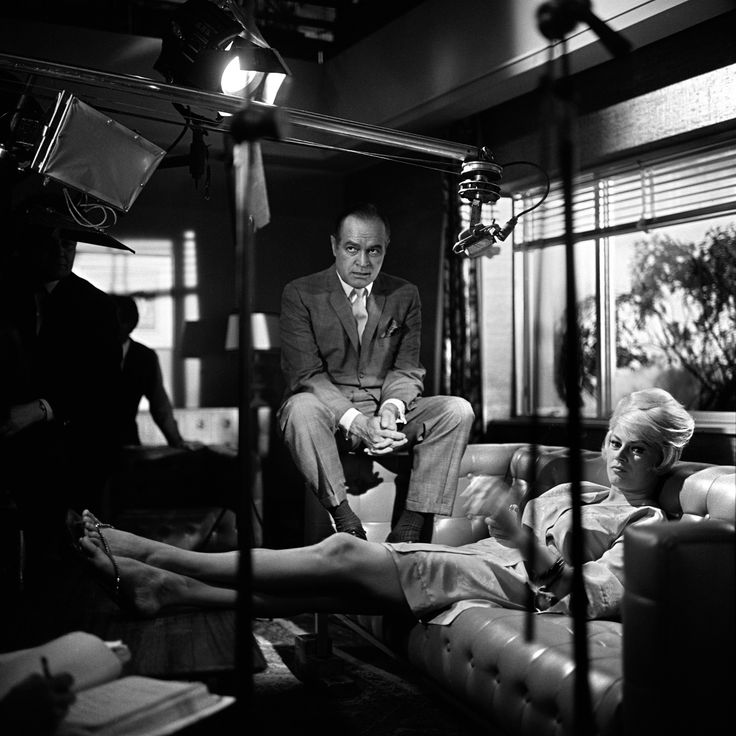} &
\includegraphics[trim={0 2cm 0 0},clip,width=0.34\columnwidth,height=0.24\columnwidth]{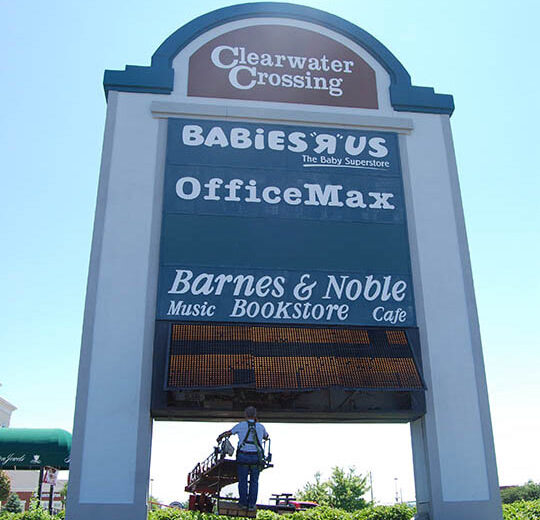} &
\includegraphics[trim={0 .1cm 0 .2cm},clip,width=0.34\columnwidth,height=0.24\columnwidth]{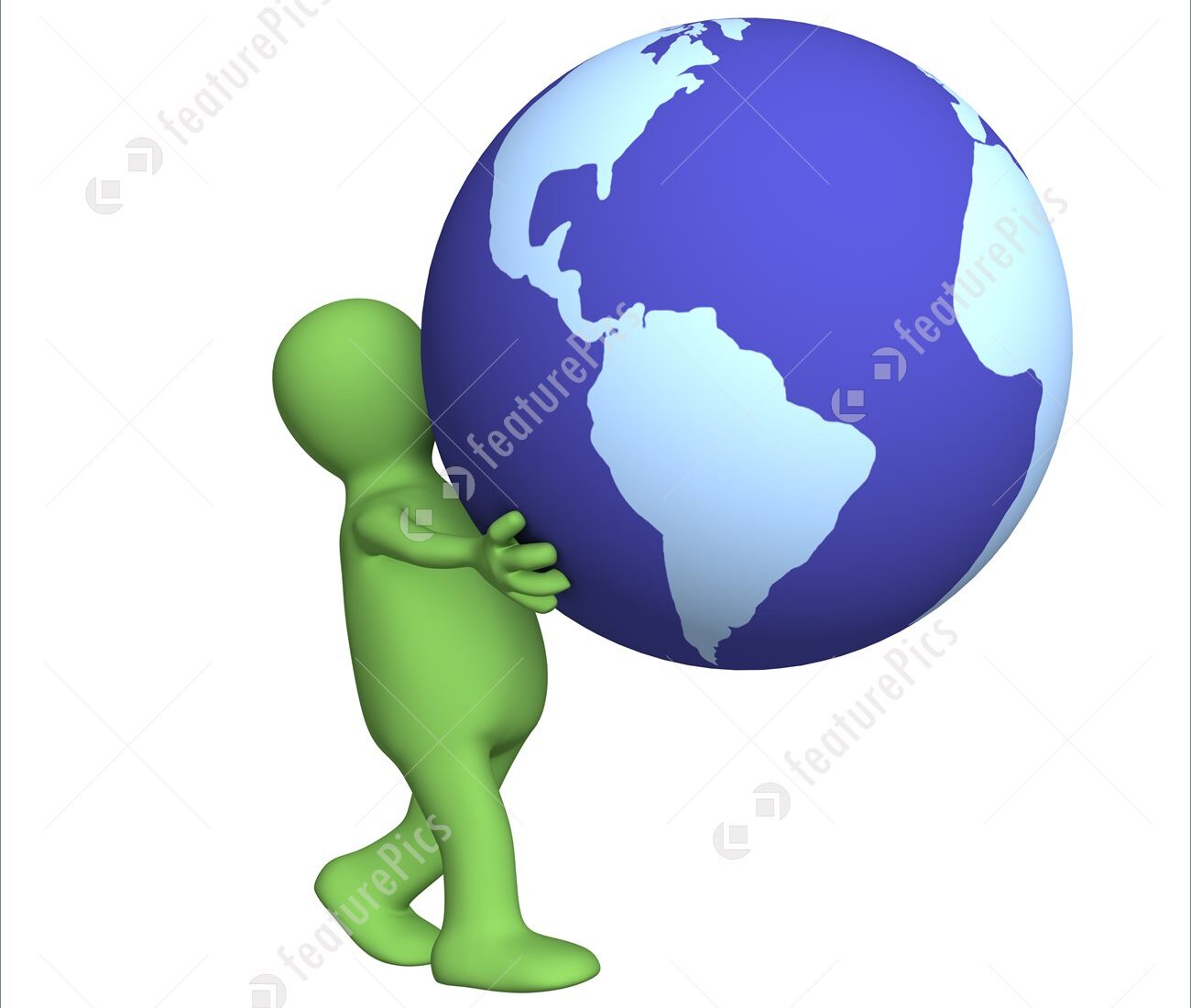} &
\includegraphics[width=0.34\columnwidth,height=0.24\columnwidth]{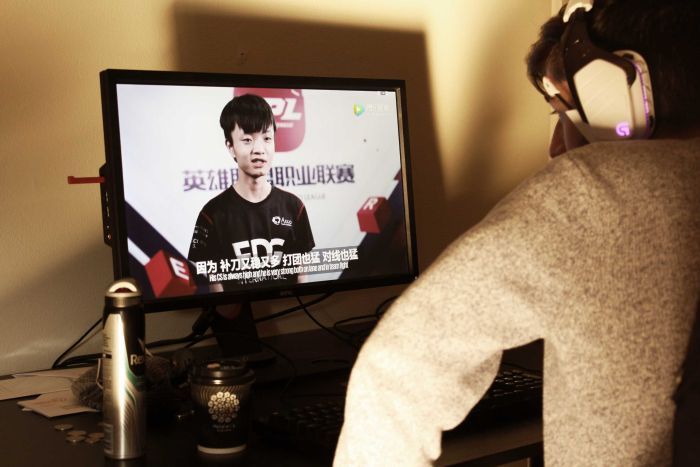} &
\includegraphics[width=0.34\columnwidth,height=0.24\columnwidth]{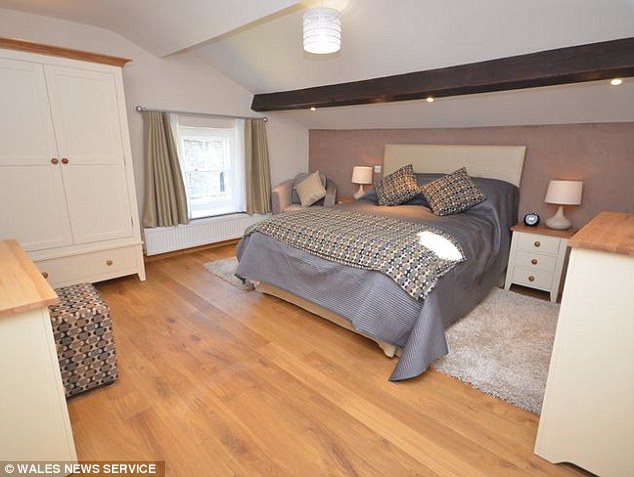} \\

% \hline
\begin{tabular}{l}Ground Truth \end{tabular}& A life in photography -- in pictures. & Photograph of the sign being repaired by brave person. 
& Globes : the green 3d person carrying in hands globe. &  The player staring intently at a computer screen.  & The - bedroom stone cottage can sleep people.\\
\hline
\begin{tabular}{l}VLP \end{tabular}&  Actors in a scene from the movie. & The sign at the entrance.  & Templates: green cartoon character holding the earth globe. & Person works on a video.  &  The master bedroom has a king - sized bed with a queen size bed.  \\
\hline
\begin{tabular}{l}
Ours; MLP +\\ GPT2 tuning
\end{tabular}& Actor sits in a hotel room. & The sign at the entrance. & 3d render of a man holding a globe. & Person, a student, watches a video on his laptop.  & The property is on the market for £ 1.\\
\hline
\begin{tabular}{l}
Ours; \\ Transformer
\end{tabular}& person sitting on a chair in a room. & a sign is seen at the entrance to the store. & stock image of a man holding the earth. & portrait of a young boy playing video game. & one of the bedrooms in the house has been converted into a living room.

\end{tabular}
%\vspace{-0.15cm}
\caption{ Uncurated results of the first five images in our test set for Conceptual Captions \cite{sharma2018conceptual}.}
%\vspace{-0.15cm}
\label{fig:img_conceptual} 
\end{figure*}

\begin{figure}
  \centering
  \footnotesize

\renewcommand{\arraystretch}{1}
\setlength{\tabcolsep}{3pt}

  \centering
  \footnotesize
\begin{tabular}{p{\imagew} p{\imagew} p{\imagew}  p{\imagew}}

\includegraphics[width=0.4\columnwidth,height=0.25\columnwidth]{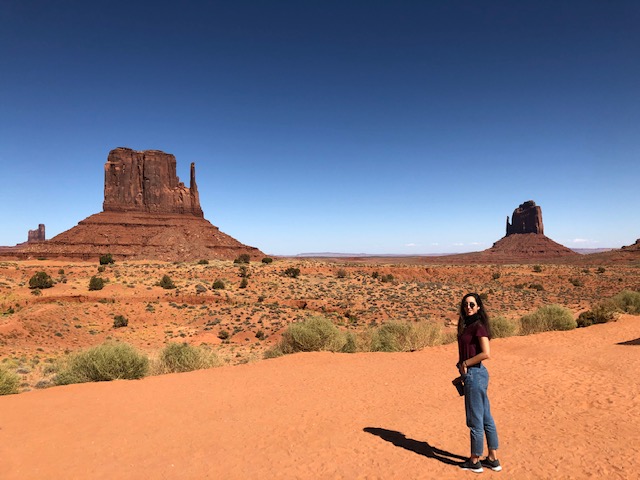} &
\includegraphics[width=0.4\columnwidth,height=0.25\columnwidth]{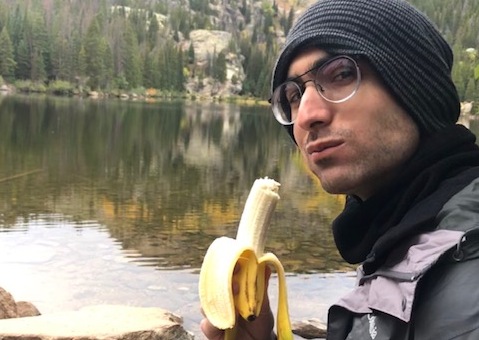} \\
A person standing in front of a rock formation in the desert. & A man holding a banana in front of a river. \\
\\
\includegraphics[width=0.4\columnwidth,height=0.25\columnwidth]{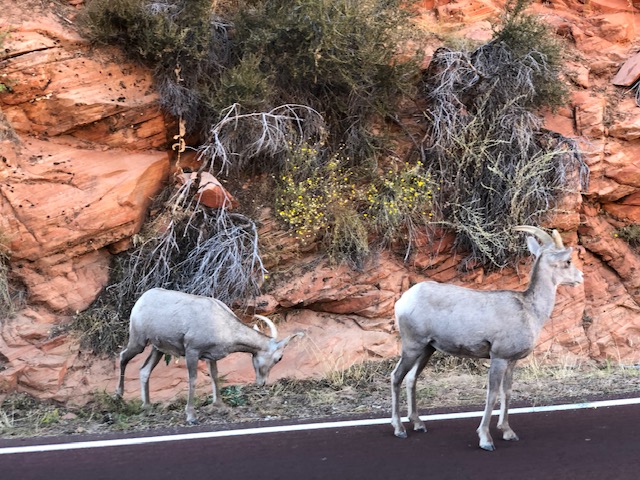} &
\includegraphics[width=0.4\columnwidth,height=0.25\columnwidth]{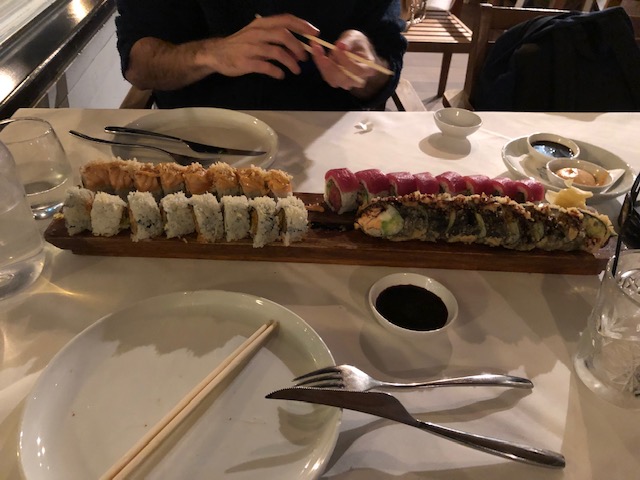} \\
Two horned goats crossing a road in the desert. & A person sitting at a table with a tray of sushi. \\
 
\end{tabular}
%\end{center}
%\vspace{-0.25cm}
\caption{Results over smartphone photos. Top: using our Conceptual Captions model. Bottom: COCO model. As demonstrated, our approach generalizes well to newly photographed images.}
%\vspace{-0.25cm} 
%\vspace*{0.4cm} 
\label{fig:real} 
\end{figure}

% IMG_2039:  A woman standing in front of a large rock formation 
% IMG_3461_2:  A man holding a banana in front of a lake .
% IMG_1900:  Two mountain goats eating grass on the side of the road .
% IMG_0277:  A table with a cake and some silverware on it .

\section{Results}
\label{sec:res} 

\paragraph{Datasets.}

We use the \textit{COCO-captions} \cite{lin2014microsoft, chen2015microsoft}, \textit{nocaps} \cite{agrawal2019nocaps} , and \textit{Conceptual Captions} \cite{sharma2018conceptual} datasets. We split the former according to the Karpathy et al.~\cite{karpathy2015deep} split, where the training set contains $120,000$ images and $5$ captions per image. Since COCO is limited to $80$ classes, the nocaps dataset is designed to measure generalization to unseen classes and concepts. It contains only validation and test sets, with the training utilizing COCO itself. The nocaps dataset is divided to three parts --- \textit{in-domain} contains images portraying only COCO classes, \textit{near-domain} contains both COCO and novel classes, and \textit{out-of-domain} consists of only novel classes. As suggested by Li et al.~\cite{li2020oscar}, we evaluate the model using only the validation set. Though some methods utilize object tags of the novel classes, we only consider the setting of no additional supervision, as we find it more applicable in practice. Therefore, we do not employ a constrained beam search \cite{anderson2016guided}.
The Conceptual Captions dataset consists of $3$M pairs of images and captions, harvested from the web and post-processed. It is considered to be more challenging than COCO due to the larger variety of styles of both the images and the captions, while not limited to specific classes. To focus on the concepts, specific entities in this dataset are replaced with general notions. For example, in Fig.~\ref{fig:teaser}, the names are replaced with "politician". For evaluation, we use the validation set, consisting of $12.5$K images, as the test set is not publicly available. Consequently, we did not use this set for validation.

\paragraph{Baselines.}
We compare our method to the state-of-the-art works of Li et al.~\cite{li2020oscar} (known as Oscar), Vision-Language Pre-training model (VLP) \cite{zhou2020unified}, and the eminent work of Anderson et al.~\cite{anderson2018bottom}, denoted BUTD. 
These models first produce visual features using an object detection network \cite{ren2015faster}. BUTD then utilizes an LSTM to generate the captions, while VLP and Oscar employ a transformer, trained similarly to BERT \cite{devlin2018bert}. Both VLP and Oscar exploit an extensive pre-trained procedure over millions of image-text pairs.
Oscar \cite{li2020oscar} also uses additional supervision compared to our setting, in the form of object tags for each image.

Our default configuration employs the transformer mapping network, without fine-tuning the language model, denoted \textbf{Ours; Transformer}. Additionally, we also evaluate our variant that utilizes the MLP mapping network, and fine-tunes the language model, denoted \textbf{Ours; MLP + GPT2 tuning}. Other configurations are evaluated in Tab.~\ref{tab:num}$(D)$.

\paragraph{Evaluation metrics.}
Similar to Li et al.~\cite{li2020oscar}, we validate our results over the COCO dataset using the common metrics BLEU \cite{papineni2002bleu}, METEOR \cite{denkowski2014meteor},  CIDEr \cite{vedantam2015cider} and SPICE \cite{anderson2016spice}, and for the nocaps dataset using CIDEr and SPICE. For the Conceptual Captions, we report the ROUGE-L \cite{lin2004automatic}, CIDEr, and SPICE, as suggested by the authors \cite{sharma2018conceptual}.

Furthermore, we measure the training time and the number of trainable parameters to validate the applicability of our method.
Reducing the training time allows to quickly obtain a new model for new data, create an ensemble of models, and decrease energy consumption. Similar to other works, we report training time in GPU hours, and the GPU model used. The number of trainable parameters is a popular measure to indicate model feasibility.

% \begin{figure}[h]
% \footnotesize 
% \begin{center}
% \begin{tabular}[t]{l @{\hskip 3\tabcolsep} p{0.35\columnwidth} @{\hskip 2\tabcolsep} p{0.35\columnwidth}}

% & {\includegraphics[trim={0 1.8cm 0 0} ,clip,width=0.35\columnwidth]{images/COCO_train2014_000000019906.jpg}} & 
% {\includegraphics[width=0.35\columnwidth]{images/COCO_train2014_000000444467.jpg}} \\
% \textbf{GPT-$2$ tuning} \vspace{1mm} \\ 
% % \parbox[t]{0mm}{\multirow{2}{*}{\rotatebox[origin=c]{90}{GPT-$2$ tuning}}}&
% Prefix & 
% com showcase motorcycle A 
% &
% vegetarian apple pies eating apples fruits both \vspace{1mm} 
% \\
% Caption & 
% a motorcycle is on display at a show.
% &
% two people wearing hats and eating apples.
% \\

% \hline
% \textbf{W/O tuning}  \vspace{1mm} \\ 
% % \parbox[t]{0mm}{\multirow{2}{*}{\rotatebox[origin=c]{90}{w/o tuning }}}
% % &
% Prefix & 
% oover eleph SniperÃÂÃÂ motorcycle synergy undeniably achieving
% &
% ockets Pier eleph NinjaÃÂÃÂ rousersNonetheless strikingly himself 
%  \vspace{1mm}  \\
% Caption &
% a motorcycle that is on display at a show.
% &
% a couple of people dressed in costumes eating apples. 
% % \bottomrule
% \end{tabular}
% \end{center}

% \caption{Prefix Interpretability. We present both the generated caption and our prefix interpretation. Upper: Ours; MLP + GPT2 tuning. Bottom: Ours; Transformer.}
% \label{fig:interpretation} 
% \end{figure}

\begin{figure*}
\footnotesize
\centering
\begin{tabular}{l @{\hskip 5\tabcolsep} p{0.12\columnwidth}  p{0.35\columnwidth}p{0.35\columnwidth}p{0.35\columnwidth}p{0.35\columnwidth}p{0.35\columnwidth}}
& &
\includegraphics[width=0.36\columnwidth,height=0.27\columnwidth]{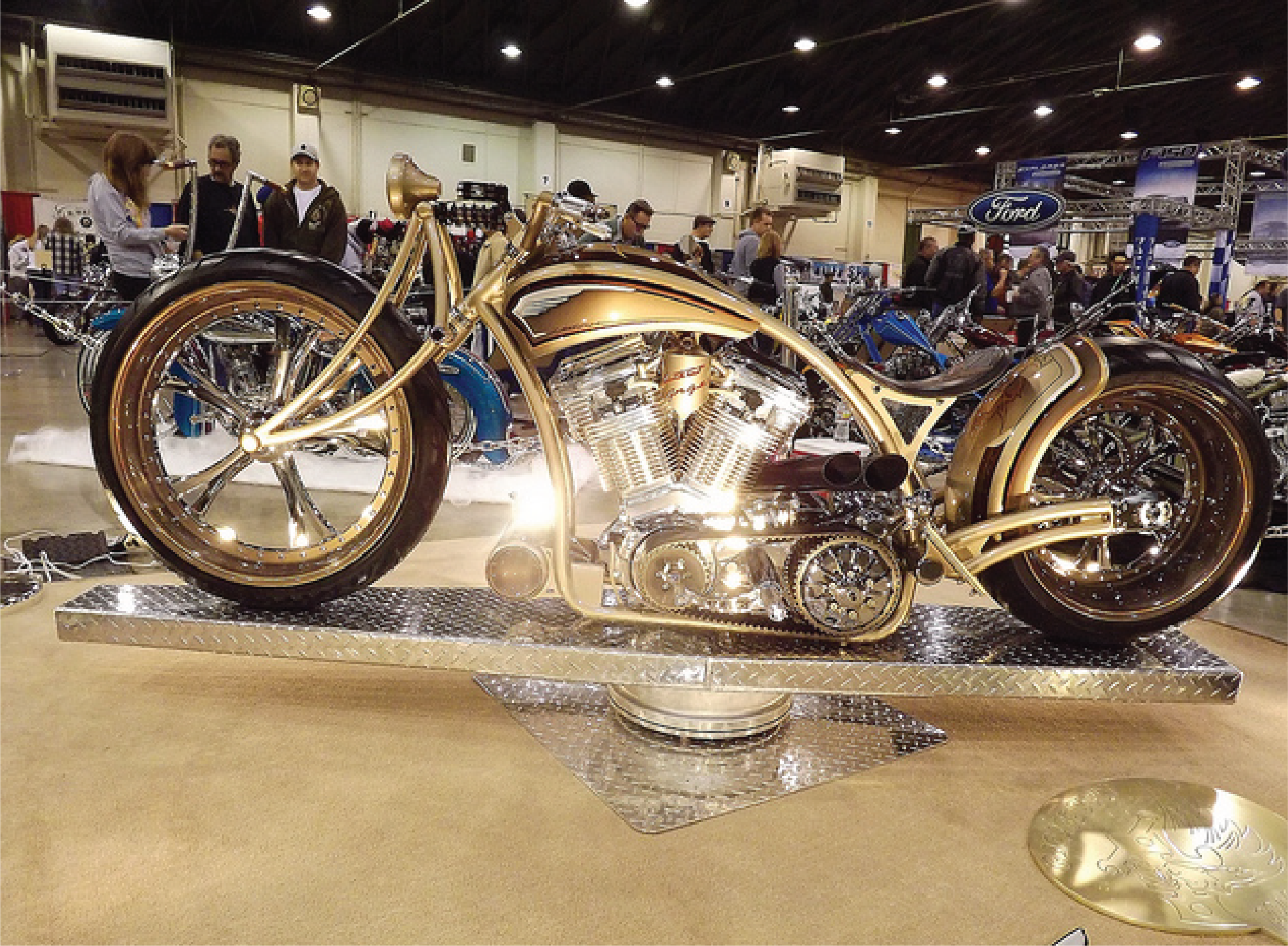} &
\includegraphics[width=0.36\columnwidth,height=0.27\columnwidth]{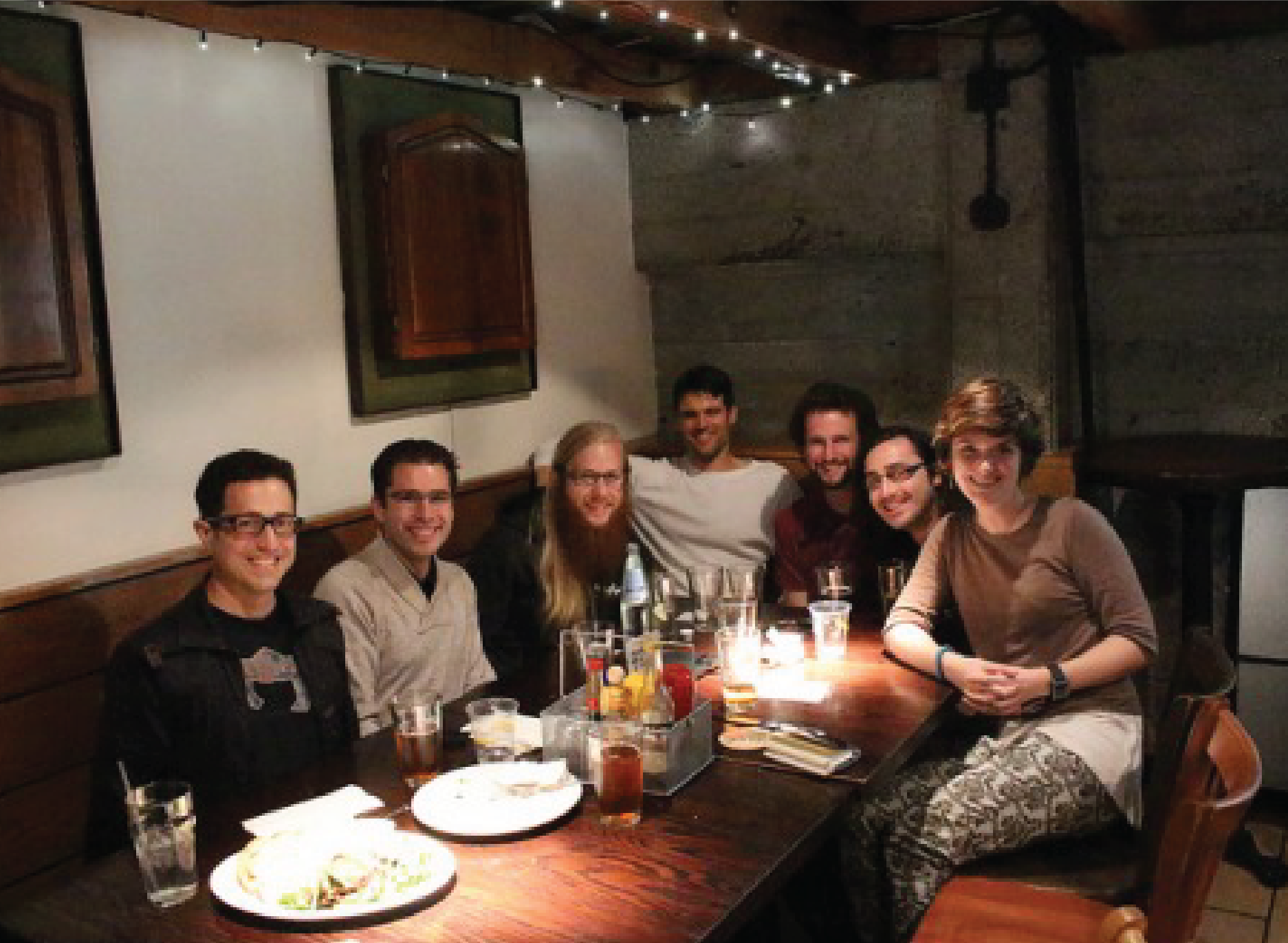} &
\includegraphics[trim={2cm 0cm 1cm 0cm},clip,width=0.36\columnwidth,height=0.27\columnwidth]{images/prefix_example-07.jpg} &
\includegraphics[width=0.36\columnwidth,height=0.27\columnwidth]{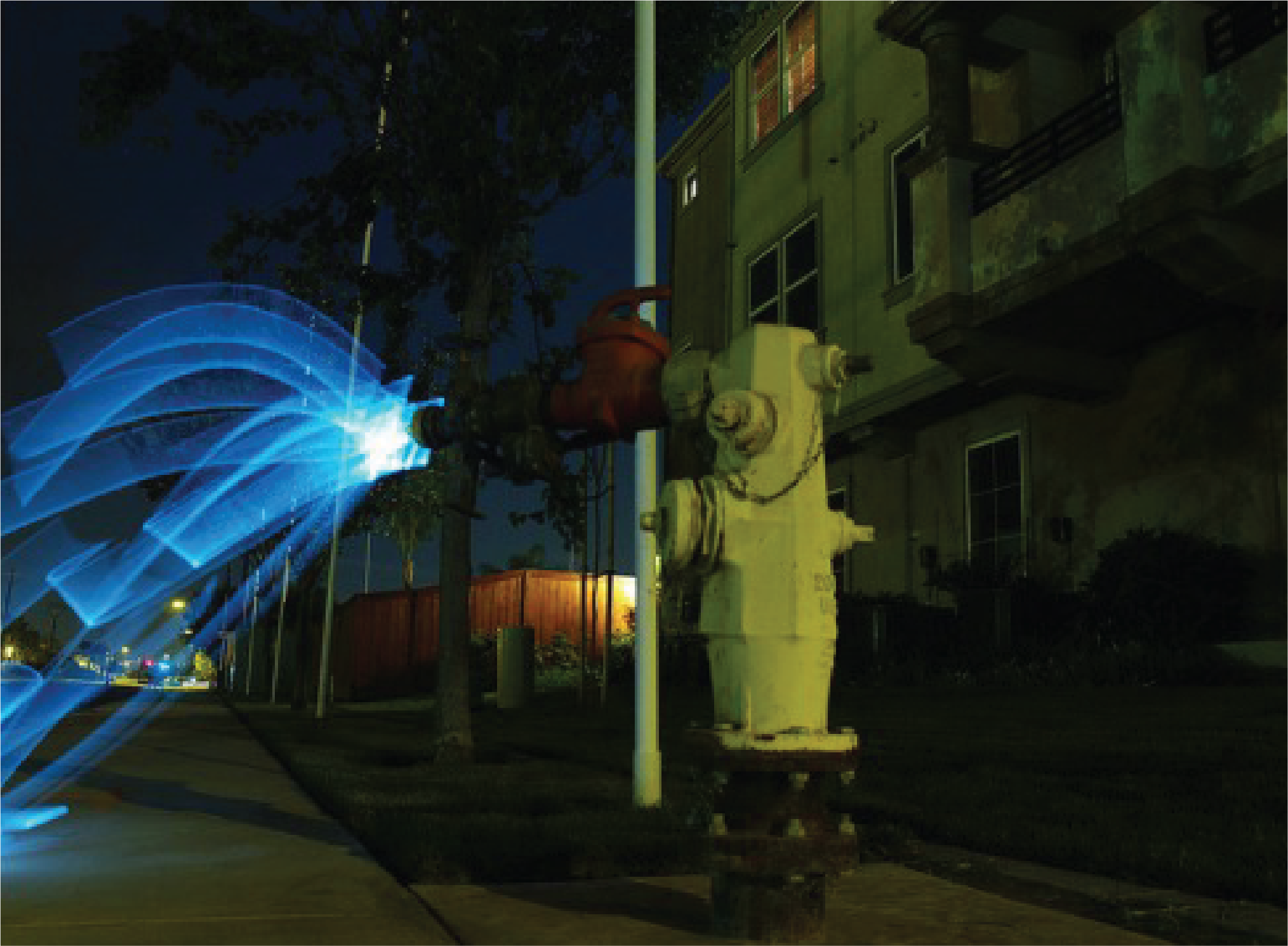} &
\includegraphics[width=0.36\columnwidth,height=0.27\columnwidth]{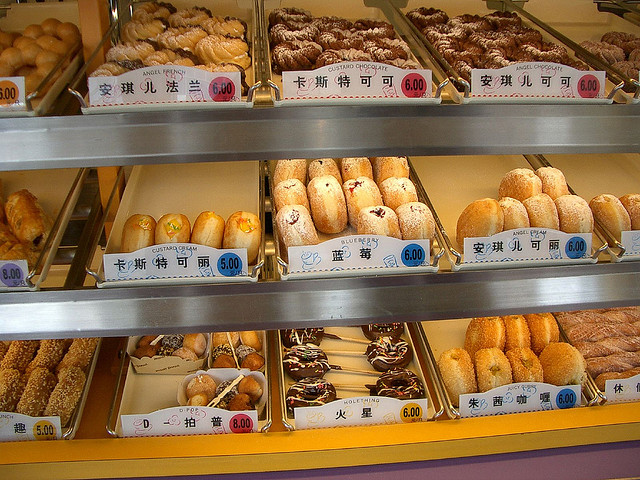} 
\\
\parbox[t]{0mm}{\multirow{2}{*}{\rotatebox[origin=c]{90}{\textbf{GPT-$2$ tuning}}}}
& Caption & a motorcycle is on display in a showroom. & a group of people sitting around a table. & a living room filled with furniture and a book shelf filled with books. & a fire hydrant is in the middle of a street. & display case filled with lots of different types of donuts. \\
& Prefix & com showcase motorcycle A ray motorcycleposed what polished Ink & blond vegetarian dishes dining expects smiling friendships group almost & tt sofa gest chair Bart books modern doorway bedroom & neon street Da alley putistan colorful nighttime & glass bakery dough displays sandwiches2 boxes Prin ten \\

\hline
\parbox[t]{0mm}{\multirow{2}{*}{\rotatebox[origin=c]{90}{\textbf{w/o tuning }}}}
& Caption & motorcycle that is on display at a show. & a a group of people sitting at a table together. & a living room with a couch and bookshelves. & a fire hydrant in front of a city street. & a display case full of different types of doughnuts.
\\
& Prefix & oover eleph SniperÃÂÃÂ motorcycle synergy undeniably achieving\textbackslash n & amic Delicious eleph SukActionCode photographers interchangeable undeniably achieving & orianclassic eleph CameroonÃÂÃÃÂroom synergy strikingly achieving\textbackslash n & ockets Pier eleph SniperÃÂÃÂ bicycl synergy undeniably achieving\textbackslash n & peanuts desserts elephbmÃÂÃÂÃ cooking nodd strikingly achieving\textbackslash n \\

\end{tabular}
%\vspace{-0.2cm}
\caption{Prefix Interpretability. We present both the generated caption and our prefix interpretation. Upper: Ours; MLP + GPT2 tuning. Bottom: Ours; Transformer.}
%\vspace{-0.4cm}
\label{fig:interpretation} 
\end{figure*}

% a living room filled with furniture and a book shelf filled with books.
% 'tt sofa gest chair Bart books modern doorway bedroom'

% a living room with a couch and bookshelves.
% orianclassic eleph CameroonÃÂÃÃÂroom synergy strikingly achieving\textbackslash n

\paragraph{Quantitative evaluation.}

Quantitative results for the challenging Conceptual Captions dataset are presented in Tab.~\ref{tab:num}$(A)$.
As can be seen, we surpass the results of VLP, while requiring orders of magnitude less training time. We note that our lightweight model, which does not fine-tune GPT-$2$, achieves an inferior result for this dataset. We hypothesize that due to the large variety of styles, a more expressive model is required than our light model, which induces a significantly lower parameter count. We compare only to VLP, as the other baselines haven't published results nor trained models for this dataset.

Tab.~\ref{tab:num}$(B)$ presents results for the nocaps dataset, where we achieve comparable results to the state-of-the-art method Oscar. As can be seen, 
Oscar achieves a slightly better SPICE score and we attain a slightly better CIDEr score. Still, our method uses a fraction of training time and trainable parameters with no additional object tags required, hence it is much more useful in practice.

Tab.~\ref{tab:num}$(C)$ present the results for the COCO dataset. Oscar reaches the best results, however, it uses additional input in the form of object tags. Our results are closed to VLP and BUTD which utilize considerably more parameters and training time. Note that the training time of VLP and Oscar does not include the pre-training step. For instance, pre-training of VLP requires training over Conceptual Captions which consumes $1200$ GPU hours.

Both Conceptual Captions and nocaps are designed to model a larger variety of visual concepts than COCO. Therefore, we conclude our method is preferable for generalizing to diverse data using a quick training procedure. This originates from utilizing the already rich semantic representations of both CLIP and GPT-$2$.

\paragraph{Qualitative evaluation.}

Visual results of the uncurated first examples in our test sets of both Conceptual Captions and COCO datasets are presented in Figs.~\ref{fig:img} and \ref{fig:img_conceptual} respectively. As can be seen, our generated captions are meaningful and depict the image successfully for both datasets. We present additional examples collected from the web in Fig.~\ref{fig:teaser}. As can be seen, our Conceptual Captions model generalizes well to arbitrary unseen images as it was trained over a sizable and diverse set of images. We also present in Fig.~\ref{fig:real} results over smartphone images,  to further demonstrate generalization to new scenarios. Moreover, our model successfully identifies uncommon objects even when trained only over COCO. For example, our method recognizes the wooden spoons or the cake with a candle better than Oscar in Fig.~\ref{fig:img}, since CLIP is pre-trained over a diverse set of images. 
However, our method still fails in some cases, such as recognizing the bicycle next to the train in Fig.~\ref{fig:img}. This is inherited from the CLIP model, which does not perceive the bicycle in the first place. We conclude that our model would benefit from improving CLIP object detection ability, but leave this direction for future work.  
For Conceptual Captions, our method mostly produces accurate captions, such as perceiving the green $3$d person in Fig.~\ref{fig:img_conceptual}. As expected, our method still suffers from data bias. For instance, it depicts the bedroom image in Fig.~\ref{fig:img_conceptual} as "The property is on the market for £ 1" after witnessing such captions of property advertising during training.

\paragraph{Language model fine-tuning.}
As described in Section.~\ref{sec:method}, fine-tuning the language model results in a much more expressive model, but that is also more susceptible to overfitting, as the amount of trainable parameters increases. As can be seen in Tab.~\ref{tab:num}, the two variants --- with and without the language model fine-tuning --- are comparable. Over the extremely complicated Conceptual Captions dataset, we get superior results with the fine-tuning. While over the popular COCO dataset, avoiding the fine-tuning achieves better results. Regarding nocaps dataset, the results are roughly equal, thus the lighter model would be preferable. We thus hypothesize that extremely elaborated datasets or ones that present a unique style require more expressiveness, and hence the more likely it is to benefit from the fine-tuning.

\paragraph{Prefix Interpretability.}

To further understand our method and results, we suggest interpreting the generated prefixes as a sequence of words. Since the prefix and word embeddings share the same latent space, they can be treated similarly. 
We define the interpretation of each of the $k$ prefix embeddings as the \textit{closest} vocabulary token, under cosine similarity.
Fig.~\ref{fig:interpretation} shows examples of images, the generated captions, and their prefix interpretations. The interpretation is meaningful when both the mapping network and GPT-$2$ are trained. In this case, the interpretation contains salient words that associate with the content of the image. For Instance, motorcycle and showcase in the first example. However, when we only train the mapping network, the interpretation becomes essentially unreadable since the network is also charged with maneuvering the fixed language model. Indeed, a considerable part of the prefix embeddings is shared across different images for the same model, as it performs the same adjustment to GPT-$2$.

\paragraph{Prefix length.}

\begin{figure}
\centering
\begin{subfigure}{\textwidth}
\begin{tikzpicture}
\tikzstyle{every node}=[font=\scriptsize]
\hspace{-.2cm}
\begin{axis}[
    axis y line*=left,
    title={},
    xlabel style={font=\footnotesize, yshift=0.2cm},
    xlabel={Prefix length},
    ylabel style={font=\footnotesize, yshift=-0.5cm, color=red},
    ylabel={BLEU@$4$},
    yticklabel style={color=red},
    xmin=.8, xmax=20.2,
    ymin=29.8, ymax=42.2,
    xtick={1,5,10, 15, 20},
    ytick={30, 34, 38, 42},
    width=.47\textwidth,
    height=.23\textwidth,
]

\addplot[mark=x,red,dashed]
  coordinates{
(1,32.192)(2,32.956)(5,35.1918)(10,37.6397)(15,38.8974)(20,40.1961)
};\label{plot_two}

 \addplot[
    color=red,
    mark=x
    ]
    coordinates {
    (1,30.)(2,30.8)(5,31.6)(10,31.5)(15,31.5)(20,31.3)
    };\label{plot_one}

% \pgfplotsset{every axis legend/.append style={
% at={(0.23,1.03)},
% anchor=south}}
% \legend{train, test}

\end{axis}
\begin{axis}[
axis y line*=right,
  ylabel near ticks,
  yticklabel pos=right,
  axis x line=none,
  yticklabel style={color=blue},
  ylabel style={font=\footnotesize, yshift=0.1cm, color=blue},
  ylabel={CIDEr},
  xmin=.8, xmax=20.2,
  ymin=99.8, ymax=140.2,
  ytick={100, 110, 120, 130, 140},
  width=.47\textwidth,
    height=.23\textwidth,
    legend pos=north west,
    legend style={fill=white, fill opacity=0.6, draw opacity=0.6,text opacity=1},
]
% \pgfplotsset{every axis legend/.append style={
% at={(0.81,1.03)},
% anchor=south}}

% \legend{CIDEr train,CIDEr test}

% \addlegendentry{\scriptsize{train}}
% \addlegendentry{\scriptsize{test}}

\addlegendimage{dashed,mark options={color=black},color=black}
\addlegendentry{train}
\addlegendimage{mark options={color=black},color=black}
\addlegendentry{test}

\addplot[mark=x,blue,dashed]
  coordinates{
(1,103.678)(2,106.92)(5,115.0527)(10,123.92543)(15,128.85635)(20, 132.393)
};\label{plot_four}
\addplot[mark=x,blue]
  coordinates{
(1,100.2)(2,103.2)(5,105.4)(10,106.1)(15,105.9)(20,105.5)
};\label{plot_three}
\end{axis}
\end{tikzpicture}
% \vspace{-0.5cm}
% \caption{GPT Tune and MLP training with varying size of the prefix length. BLEU@$4$ and CIDEr scores over COCO-captions dataset as a function of the prefix length.
% % \amirc{Large MLPs are not feasible: For prefix size of 40, the mapping network contains: $(512 \times 20 \times 768) + (20 \times 768 \times 40 \times 768)  > 450M$ parameters}}
%     \label{fig:prefix_legnth}
\subcaption{MLP mapping network with fine-tuning of the language model.}
\label{fig:prefix_legnth_mlp}
\end{subfigure}
\begin{subfigure}{\textwidth}
\vspace{3mm}
\begin{tikzpicture}
\tikzstyle{every node}=[font=\scriptsize]
\hspace{-.2cm}
\begin{axis}[
    axis y line*=left,
    title={},
    xlabel style={font=\footnotesize, yshift=0.2cm},
    xlabel={Prefix length},
    ylabel style={font=\footnotesize, yshift=-0.5cm, color=red},
    ylabel={BLEU@$4$},
    yticklabel style={color=red},
    xmin=.8, xmax=80.2,
    ymin=25.8, ymax=42.2,
    xtick={1,5,10, 20, 40, 80},
    ytick={26, 30, 34, 38, 42},
    width=.47\textwidth,
    height=.23\textwidth,
]

\addplot[mark=x,red,dashed]
  coordinates{
(1,26.746)(2,28.296)(5,29.275)(10,33.518)(20,35.530)(40,40.015)(80,40.9455)
};\label{plot_two}

 \addplot[
    color=red,
    mark=x
    ]
    coordinates {
    (1,28.746)(2,31.85)(5,32.42)(10,33.53)(20,32.9)(40,33.57)(80,34.244)
    };\label{plot_one}

% \pgfplotsset{every axis legend/.append style={
% at={(0.23,1.03)},
% anchor=south}}
% \legend{train, test}

\end{axis}
\begin{axis}[
axis y line*=right,
  ylabel near ticks,
  yticklabel pos=right,
  axis x line=none,
  yticklabel style={color=blue},
  ylabel style={font=\footnotesize, yshift=0.1cm, color=blue},
  ylabel={CIDEr},
  xmin=.8, xmax=80.2,
  ymin=89.8, ymax=140.2,
  ytick={90, 100, 110, 120, 130, 140},
  width=.47\textwidth,
    height=.23\textwidth,
    legend pos=north west,
    legend style={fill=white, fill opacity=0.6, draw opacity=0.6,text opacity=1},
]
% \pgfplotsset{every axis legend/.append style={
% at={(0.81,1.03)},
% anchor=south}}

% \legend{CIDEr train,CIDEr test}

% \addlegendentry{\scriptsize{train}}
% \addlegendentry{\scriptsize{test}}

\addlegendimage{dashed,mark options={color=black},color=black}
\addlegendentry{train}
\addlegendimage{mark options={color=black},color=black}
\addlegendentry{test}

\addplot[mark=x,blue,dashed]
  coordinates{
(1,90.0955)(2,95.14)(5,100.6905)(10,111.6430)(20,118.827)(40,132.8)(80,136.500)
};\label{plot_four}

\addplot[mark=x,blue]
  coordinates{
(1,101.04)(2,105.38)(5,110.168)(10,113.08)(20,111.59)(40,112.89)(80,115.069)
};\label{plot_three}
\end{axis}
\end{tikzpicture}
\subcaption{Transformer mapping network with frozen language model.}
\label{fig:prefix_legnth_transformer}
\end{subfigure}
%\vspace{-0.15cm}
\caption{Effect of the prefix length on the captioning performance over the COCO-captions dataset. For each prefix length, we report the BLEU@4 (red) and CIDERr (blue) scores over the test and train (dashed line) sets.}
%\vspace{-0.3cm}
 \label{fig:prefix_legnth}
\end{figure}
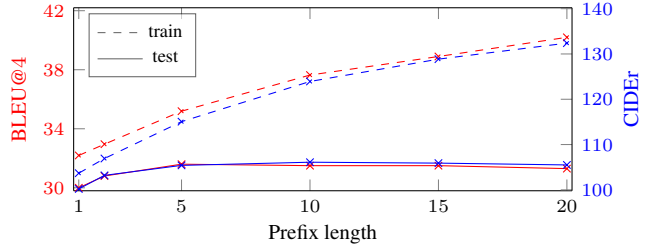
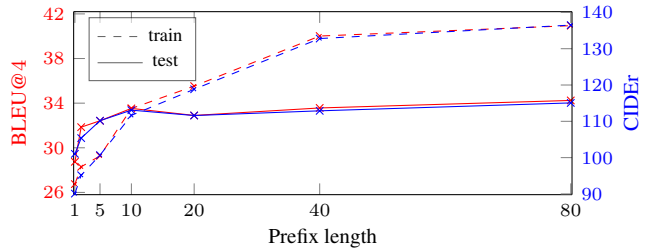

Li and Liang \cite{li2021prefix} showed that increasing the size of the prefix length, up to a certain value, improves the performance of the model in an underlying task. Moreover, the saturation length might differ between tasks. For the image captioning task, we conduct an ablation over the prefix lengths using the COCO dataset over two configurations of our method:
\textbf{Ours; Transformer} and \textbf{Ours; MLP + GPT2 tuning}. 
The results are summarized in Fig.~\ref{fig:prefix_legnth}. For each prefix size and configuration, we train the network for $5$ epochs and report the BLEU@$4$ and CIDEr scores over the test and train sets.

As can be seen in Fig.~\ref{fig:prefix_legnth_mlp}, increasing the prefix size while allowing tuning of the language model results in overfitting to the training set, due to the large number of trainable parameters. However, when the language model is frozen, we experience improvement for both the training and test evaluations, as can be seen in Fig.~\ref{fig:prefix_legnth_transformer}. Naturally, extremely small prefix length yields inferior results as the model is not expressive enough. In addition, we point out that the MLP architecture is inherently more limited as it is not scalable for a long prefix. For example, a prefix size of $40$ implies a network with over $450$M parameters, which is unfeasible for our single GPU setting. The transformer architecture allows increasing the prefix size with only marginal increment to the number of the parameters, but only up to $80$ --- due to the quadratic memory cost of the attention mechanism.

\paragraph{Mapping network.}
An ablation study for the mapping network architecture is shown in Tab.~\ref{tab:num}$(C)$,$(D)$. As can be seen, with language model fine-tuning, the MLP achieves better results. However, the transformer is superior when the language model is frozen. We conclude that when employing the fine-tuning of the language model, the expressive power of the transformer architecture is unnecessary.

\captionsetup[subfigure]{font=small, justification=justified,singlelinecheck=false}

\paragraph{Implementation details.}

We used the prefix length of $K=10$ for the MLP mapping networks, where the MLP contains a single hidden layer. For the transformer mapping network, we set the CLIP embedding to $K=10$ constants tokens and use $8$ multi-head self-attention layers with $8$ heads each. We train for $10$ epochs using a batch size of $40$. For optimization, we use AdamW \cite{Kingma2015AdamAM} with weight decay fix as introduced by Loshchilov et al.~\cite{loshchilov2017decoupled}, with a learning rate of $2e^{-5}$ and $5000$ warm-up steps. For GPT-$2$ we employ the implementation of Wolf et al.\cite{wolf-etal-2020-transformers}.

\section{Conclusion}

Overall, our CLIP-based image-captioning method is simple to use, doesn't require any additional annotations, and is faster to train. Even though we propose a simpler model, it demonstrates more merit as the dataset becomes richer and more diverse. We consider our approach as part of a new image captioning paradigm, concentrating on leveraging existing models, while only training a minimal mapping network. This approach essentially learns to adapt existing semantic understanding of the pre-trained models to the style of the target dataset, instead of learning new semantic entities. We believe the utilization of these powerful pre-trained models would gain traction in the near future. Therefore, the understanding of how to harness these components is of great interest. For future work, we plan to incorporate pre-trained models (e.g., CLIP), to other challenging tasks, such as visual question answering or image to $3$D translation, through the utilization of mapping networks.

%%%%%%%%% REFERENCES
{\small
\bibliographystyle{ieee_fullname}
\bibliography{egbib}

\begin{thebibliography}{10}\itemsep=-1pt

\bibitem{agrawal2019nocaps}
Harsh Agrawal, Karan Desai, Yufei Wang, Xinlei Chen, Rishabh Jain, Mark
  Johnson, Dhruv Batra, Devi Parikh, Stefan Lee, and Peter Anderson.
\newblock nocaps: novel object captioning at scale.
\newblock In {\em Proceedings of the IEEE/CVF International Conference on
  Computer Vision}, pages 8948--8957, 2019.

\bibitem{anderson2016guided}
Peter Anderson, Basura Fernando, Mark Johnson, and Stephen Gould.
\newblock Guided open vocabulary image captioning with constrained beam search.
\newblock {\em arXiv preprint arXiv:1612.00576}, 2016.

\bibitem{anderson2016spice}
Peter Anderson, Basura Fernando, Mark Johnson, and Stephen Gould.
\newblock Spice: Semantic propositional image caption evaluation.
\newblock In {\em European conference on computer vision}, pages 382--398.
  Springer, 2016.

\bibitem{anderson2018bottom}
Peter Anderson, Xiaodong He, Chris Buehler, Damien Teney, Mark Johnson, Stephen
  Gould, and Lei Zhang.
\newblock Bottom-up and top-down attention for image captioning and visual
  question answering.
\newblock In {\em Proceedings of the IEEE conference on computer vision and
  pattern recognition}, pages 6077--6086, 2018.

\bibitem{bau2021paint}
David Bau, Alex Andonian, Audrey Cui, YeonHwan Park, Ali Jahanian, Aude Oliva,
  and Antonio Torralba.
\newblock Paint by word.
\newblock {\em arXiv preprint arXiv:2103.10951}, 2021.

\bibitem{chen2017sca}
Long Chen, Hanwang Zhang, Jun Xiao, Liqiang Nie, Jian Shao, Wei Liu, and
  Tat-Seng Chua.
\newblock Sca-cnn: Spatial and channel-wise attention in convolutional networks
  for image captioning.
\newblock In {\em Proceedings of the IEEE conference on computer vision and
  pattern recognition}, pages 5659--5667, 2017.

\bibitem{chen2015microsoft}
Xinlei Chen, Hao Fang, Tsung-Yi Lin, Ramakrishna Vedantam, Saurabh Gupta, Piotr
  Doll{\'a}r, and C~Lawrence Zitnick.
\newblock Microsoft coco captions: Data collection and evaluation server.
\newblock {\em arXiv preprint arXiv:1504.00325}, 2015.

\bibitem{chen2018regularizing}
Xinpeng Chen, Lin Ma, Wenhao Jiang, Jian Yao, and Wei Liu.
\newblock Regularizing rnns for caption generation by reconstructing the past
  with the present.
\newblock In {\em Proceedings of the IEEE Conference on Computer Vision and
  Pattern Recognition}, pages 7995--8003, 2018.

\bibitem{chen2014learning}
Xinlei Chen and C~Lawrence Zitnick.
\newblock Learning a recurrent visual representation for image caption
  generation.
\newblock {\em arXiv preprint arXiv:1411.5654}, 2014.

\bibitem{denkowski2014meteor}
Michael Denkowski and Alon Lavie.
\newblock Meteor universal: Language specific translation evaluation for any
  target language.
\newblock In {\em Proceedings of the ninth workshop on statistical machine
  translation}, pages 376--380, 2014.

\bibitem{devlin2018bert}
Jacob Devlin, Ming-Wei Chang, Kenton Lee, and Kristina Toutanova.
\newblock Bert: Pre-training of deep bidirectional transformers for language
  understanding.
\newblock {\em arXiv preprint arXiv:1810.04805}, 2018.

\bibitem{dosovitskiy2020image}
Alexey Dosovitskiy, Lucas Beyer, Alexander Kolesnikov, Dirk Weissenborn,
  Xiaohua Zhai, Thomas Unterthiner, Mostafa Dehghani, Matthias Minderer, Georg
  Heigold, Sylvain Gelly, et~al.
\newblock An image is worth 16x16 words: Transformers for image recognition at
  scale.
\newblock {\em arXiv preprint arXiv:2010.11929}, 2020.

\bibitem{fang2015captions}
Hao Fang, Saurabh Gupta, Forrest Iandola, Rupesh~K Srivastava, Li Deng, Piotr
  Doll{\'a}r, Jianfeng Gao, Xiaodong He, Margaret Mitchell, John~C Platt,
  et~al.
\newblock From captions to visual concepts and back.
\newblock In {\em Proceedings of the IEEE conference on computer vision and
  pattern recognition}, pages 1473--1482, 2015.

\bibitem{gal2021stylegan}
Rinon Gal, Or Patashnik, Haggai Maron, Gal Chechik, and Daniel Cohen-Or.
\newblock Stylegan-nada: Clip-guided domain adaptation of image generators.
\newblock {\em arXiv preprint arXiv:2108.00946}, 2021.

\bibitem{gao2019self}
Junlong Gao, Shiqi Wang, Shanshe Wang, Siwei Ma, and Wen Gao.
\newblock Self-critical n-step training for image captioning.
\newblock In {\em Proceedings of the IEEE/CVF Conference on Computer Vision and
  Pattern Recognition}, pages 6300--6308, 2019.

\bibitem{herdade2019image}
Simao Herdade, Armin Kappeler, Kofi Boakye, and Joao Soares.
\newblock Image captioning: Transforming objects into words.
\newblock {\em arXiv preprint arXiv:1906.05963}, 2019.

\bibitem{karpathy2015deep}
Andrej Karpathy and Li Fei-Fei.
\newblock Deep visual-semantic alignments for generating image descriptions.
\newblock In {\em Proceedings of the IEEE conference on computer vision and
  pattern recognition}, pages 3128--3137, 2015.

\bibitem{Kingma2015AdamAM}
Diederik~P. Kingma and Jimmy Ba.
\newblock Adam: A method for stochastic optimization.
\newblock {\em CoRR}, abs/1412.6980, 2015.

\bibitem{li2020oscar}
Xiujun Li, Xi Yin, Chunyuan Li, Pengchuan Zhang, Xiaowei Hu, Lei Zhang, Lijuan
  Wang, Houdong Hu, Li Dong, Furu Wei, et~al.
\newblock Oscar: Object-semantics aligned pre-training for vision-language
  tasks.
\newblock In {\em European Conference on Computer Vision}, pages 121--137.
  Springer, 2020.

\bibitem{li2021prefix}
Xiang~Lisa Li and Percy Liang.
\newblock Prefix-tuning: Optimizing continuous prompts for generation.
\newblock {\em arXiv preprint arXiv:2101.00190}, 2021.

\bibitem{lin2004automatic}
Chin-Yew Lin and Franz~Josef Och.
\newblock Automatic evaluation of machine translation quality using longest
  common subsequence and skip-bigram statistics.
\newblock In {\em Proceedings of the 42nd Annual Meeting of the Association for
  Computational Linguistics (ACL-04)}, pages 605--612, 2004.

\bibitem{lin2014microsoft}
Tsung-Yi Lin, Michael Maire, Serge Belongie, James Hays, Pietro Perona, Deva
  Ramanan, Piotr Doll{\'a}r, and C~Lawrence Zitnick.
\newblock Microsoft coco: Common objects in context.
\newblock In {\em European conference on computer vision}, pages 740--755.
  Springer, 2014.

\bibitem{liu2021cptr}
Wei Liu, Sihan Chen, Longteng Guo, Xinxin Zhu, and Jing Liu.
\newblock Cptr: Full transformer network for image captioning.
\newblock {\em arXiv preprint arXiv:2101.10804}, 2021.

\bibitem{loshchilov2017decoupled}
Ilya Loshchilov and Frank Hutter.
\newblock Decoupled weight decay regularization.
\newblock {\em arXiv preprint arXiv:1711.05101}, 2017.

\bibitem{lu2019vilbert}
Jiasen Lu, Dhruv Batra, Devi Parikh, and Stefan Lee.
\newblock Vilbert: Pretraining task-agnostic visiolinguistic representations
  for vision-and-language tasks.
\newblock {\em arXiv preprint arXiv:1908.02265}, 2019.

\bibitem{luo2021dual}
Yunpeng Luo, Jiayi Ji, Xiaoshuai Sun, Liujuan Cao, Yongjian Wu, Feiyue Huang,
  Chia-Wen Lin, and Rongrong Ji.
\newblock Dual-level collaborative transformer for image captioning.
\newblock {\em arXiv preprint arXiv:2101.06462}, 2021.

\bibitem{papineni2002bleu}
Kishore Papineni, Salim Roukos, Todd Ward, and Wei-Jing Zhu.
\newblock Bleu: a method for automatic evaluation of machine translation.
\newblock In {\em Proceedings of the 40th annual meeting of the Association for
  Computational Linguistics}, pages 311--318, 2002.

\bibitem{patashnik2021styleclip}
Or Patashnik, Zongze Wu, Eli Shechtman, Daniel Cohen-Or, and Dani Lischinski.
\newblock Styleclip: Text-driven manipulation of stylegan imagery.
\newblock In {\em Proceedings of the IEEE/CVF International Conference on
  Computer Vision}, pages 2085--2094, 2021.

\bibitem{radford2021learning}
Alec Radford, Jong~Wook Kim, Chris Hallacy, Aditya Ramesh, Gabriel Goh,
  Sandhini Agarwal, Girish Sastry, Amanda Askell, Pamela Mishkin, Jack Clark,
  et~al.
\newblock Learning transferable visual models from natural language
  supervision.
\newblock {\em arXiv preprint arXiv:2103.00020}, 2021.

\bibitem{radford2019language}
Alec Radford, Jeffrey Wu, Rewon Child, David Luan, Dario Amodei, Ilya
  Sutskever, et~al.
\newblock Language models are unsupervised multitask learners.
\newblock {\em OpenAI blog}, 1(8):9, 2019.

\bibitem{ren2015faster}
Shaoqing Ren, Kaiming He, Ross Girshick, and Jian Sun.
\newblock Faster r-cnn: Towards real-time object detection with region proposal
  networks.
\newblock {\em Advances in neural information processing systems}, 28:91--99,
  2015.

\bibitem{rennie2017self}
Steven~J Rennie, Etienne Marcheret, Youssef Mroueh, Jerret Ross, and Vaibhava
  Goel.
\newblock Self-critical sequence training for image captioning.
\newblock In {\em Proceedings of the IEEE conference on computer vision and
  pattern recognition}, pages 7008--7024, 2017.

\bibitem{sharma2018conceptual}
Piyush Sharma, Nan Ding, Sebastian Goodman, and Radu Soricut.
\newblock Conceptual captions: A cleaned, hypernymed, image alt-text dataset
  for automatic image captioning.
\newblock In {\em Proceedings of the 56th Annual Meeting of the Association for
  Computational Linguistics (Volume 1: Long Papers)}, pages 2556--2565, 2018.

\bibitem{stefanini2021show}
Matteo Stefanini, Marcella Cornia, Lorenzo Baraldi, Silvia Cascianelli,
  Giuseppe Fiameni, and Rita Cucchiara.
\newblock From show to tell: A survey on image captioning.
\newblock {\em arXiv preprint arXiv:2107.06912}, 2021.

\bibitem{tan2019lxmert}
Hao Tan and Mohit Bansal.
\newblock Lxmert: Learning cross-modality encoder representations from
  transformers.
\newblock {\em arXiv preprint arXiv:1908.07490}, 2019.

\bibitem{vaswani2017attention}
Ashish Vaswani, Noam Shazeer, Niki Parmar, Jakob Uszkoreit, Llion Jones,
  Aidan~N Gomez, {\L}ukasz Kaiser, and Illia Polosukhin.
\newblock Attention is all you need.
\newblock In {\em Advances in neural information processing systems}, pages
  5998--6008, 2017.

\bibitem{vedantam2015cider}
Ramakrishna Vedantam, C Lawrence~Zitnick, and Devi Parikh.
\newblock Cider: Consensus-based image description evaluation.
\newblock In {\em Proceedings of the IEEE conference on computer vision and
  pattern recognition}, pages 4566--4575, 2015.

\bibitem{vinyals2015show}
Oriol Vinyals, Alexander Toshev, Samy Bengio, and Dumitru Erhan.
\newblock Show and tell: A neural image caption generator.
\newblock In {\em Proceedings of the IEEE conference on computer vision and
  pattern recognition}, pages 3156--3164, 2015.

\bibitem{wang2017skeleton}
Yufei Wang, Zhe Lin, Xiaohui Shen, Scott Cohen, and Garrison~W Cottrell.
\newblock Skeleton key: Image captioning by skeleton-attribute decomposition.
\newblock In {\em Proceedings of the IEEE conference on computer vision and
  pattern recognition}, pages 7272--7281, 2017.

\bibitem{wang2021simvlm}
Zirui Wang, Jiahui Yu, Adams~Wei Yu, Zihang Dai, Yulia Tsvetkov, and Yuan Cao.
\newblock Simvlm: Simple visual language model pretraining with weak
  supervision.
\newblock {\em arXiv preprint arXiv:2108.10904}, 2021.

\bibitem{wolf-etal-2020-transformers}
Thomas Wolf, Lysandre Debut, Victor Sanh, Julien Chaumond, Clement Delangue,
  Anthony Moi, Pierric Cistac, Tim Rault, Rémi Louf, Morgan Funtowicz, Joe
  Davison, Sam Shleifer, Patrick von Platen, Clara Ma, Yacine Jernite, Julien
  Plu, Canwen Xu, Teven~Le Scao, Sylvain Gugger, Mariama Drame, Quentin Lhoest,
  and Alexander~M. Rush.
\newblock Transformers: State-of-the-art natural language processing.
\newblock In {\em Proceedings of the 2020 Conference on Empirical Methods in
  Natural Language Processing: System Demonstrations}, pages 38--45, Online,
  Oct. 2020. Association for Computational Linguistics.

\bibitem{xu2015show}
Kelvin Xu, Jimmy Ba, Ryan Kiros, Kyunghyun Cho, Aaron Courville, Ruslan
  Salakhudinov, Rich Zemel, and Yoshua Bengio.
\newblock Show, attend and tell: Neural image caption generation with visual
  attention.
\newblock In {\em International conference on machine learning}, pages
  2048--2057. PMLR, 2015.

\bibitem{yang2019learning}
Xu Yang, Hanwang Zhang, and Jianfei Cai.
\newblock Learning to collocate neural modules for image captioning.
\newblock In {\em Proceedings of the IEEE/CVF International Conference on
  Computer Vision}, pages 4250--4260, 2019.

\bibitem{yao2018exploring}
Ting Yao, Yingwei Pan, Yehao Li, and Tao Mei.
\newblock Exploring visual relationship for image captioning.
\newblock In {\em Proceedings of the European conference on computer vision
  (ECCV)}, pages 684--699, 2018.

\bibitem{zhang2017actor}
Li Zhang, Flood Sung, Feng Liu, Tao Xiang, Shaogang Gong, Yongxin Yang, and
  Timothy~M Hospedales.
\newblock Actor-critic sequence training for image captioning.
\newblock {\em arXiv preprint arXiv:1706.09601}, 2017.

\bibitem{zhang2021vinvl}
Pengchuan Zhang, Xiujun Li, Xiaowei Hu, Jianwei Yang, Lei Zhang, Lijuan Wang,
  Yejin Choi, and Jianfeng Gao.
\newblock Vinvl: Revisiting visual representations in vision-language models.
\newblock In {\em Proceedings of the IEEE/CVF Conference on Computer Vision and
  Pattern Recognition}, pages 5579--5588, 2021.

\bibitem{zhou2020unified}
Luowei Zhou, Hamid Palangi, Lei Zhang, Houdong Hu, Jason Corso, and Jianfeng
  Gao.
\newblock Unified vision-language pre-training for image captioning and vqa.
\newblock In {\em Proceedings of the AAAI Conference on Artificial
  Intelligence}, volume~34, pages 13041--13049, 2020.

\end{thebibliography}
}

\end{document}